\documentclass[letterpaper]{article} % DO NOT CHANGE THIS
\usepackage{aaai23}  % DO NOT CHANGE THIS
\usepackage{times}  % DO NOT CHANGE THIS
\usepackage{helvet}  % DO NOT CHANGE THIS
\usepackage{courier}  % DO NOT CHANGE THIS
\usepackage[hyphens]{url}  % DO NOT CHANGE THIS
\usepackage{graphicx} % DO NOT CHANGE THIS
\usepackage{natbib}  % DO NOT CHANGE THIS AND DO NOT ADD ANY OPTIONS TO IT
\usepackage{caption} % DO NOT CHANGE THIS AND DO NOT ADD ANY OPTIONS TO IT
\usepackage{algorithm}
\usepackage{algorithmic}

\usepackage{newfloat}
\usepackage{listings}
\DeclareCaptionStyle{ruled}{labelfont=normalfont,labelsep=colon,strut=off} % DO NOT CHANGE THIS
\floatstyle{ruled}
\newfloat{listing}{tb}{lst}{}
\floatname{listing}{Listing}
\usepackage{color}
\definecolor{blue}{RGB}{60,132,196}
\definecolor{red}{RGB}{207,78,56}
\definecolor{gray}{RGB}{146,146,161}
\definecolor{green4}{RGB}{46, 139, 87}
\usepackage[pagebackref,breaklinks,colorlinks,citecolor=green4]{hyperref}
\title{Deconstructed Generation-Based Zero-Shot Model}
\usepackage{marvosym}
\usepackage{ifsym}
\author {
   Dubing Chen\textsuperscript{\rm 1},
   Yuming Shen\textsuperscript{\rm 2},
   Haofeng Zhang\textsuperscript{\rm $1$\Letter},
   Philip H.S. Torr\textsuperscript{\rm 2}
}
\affiliations {
    \textsuperscript{\rm 1} Nanjing University of Science and Technology\\
    \textsuperscript{\rm 2} University of Oxford\\
  \{db.chen, zhanghf\}@njust.edu.cn,  ymcidence@gmail.com, philip.torr@eng.ox.ac.uk
}

\usepackage{bibentry}
\usepackage{amsmath}
\usepackage{amsthm}
\usepackage{multirow}
\usepackage{booktabs}
\usepackage[switch]{lineno}
\usepackage{subfigure}
\usepackage{dsfont}
\usepackage{amssymb}
\newtheorem{theorem}{Proposition}[section]
\usepackage{listings}
\newcommand{\ie}{\textit{i.e.}}
\newcommand{\eg}{\textit{e.g.}}
\newcommand{\wrt}{\textit{w.r.t.}}

\makeatletter
\newcommand*\bigcdot{\mathpalette\bigcdot@{.5}}
\newcommand*\bigcdot@[2]{\mathbin{\vcenter{\hbox{\scalebox{#2}{$\m@th#1\bullet$}}}}}
\makeatother
\begin{document}

\maketitle

\begin{abstract}
Recent research on Generalized Zero-Shot Learning (GZSL) has focused primarily on generation-based methods. However, current literature has overlooked the fundamental principles of these methods and has made limited progress in a complex manner. In this paper, we aim to deconstruct the generator-classifier framework and provide guidance for its improvement and extension. We begin by breaking down the generator-learned unseen class distribution into class-level and instance-level distributions. Through our analysis of the role of these two types of distributions in solving the GZSL problem, we generalize the focus of the generation-based approach, emphasizing the importance of (i) attribute generalization in generator learning and (ii) independent classifier learning with partially biased data. We present a simple method based on this analysis that outperforms SotAs on four public GZSL datasets, demonstrating the validity of our deconstruction. Furthermore, our proposed method remains effective even without a generative model, representing a step towards simplifying the generator-classifier structure. Our code is available at \url{https://github.com/cdb342/DGZ}.
\end{abstract}

\section{Introduction}
Big data fuels the progress of deep learning, but obtaining specific data can sometimes prove difficult. In cases where specific data is not available, Zero-Shot Learning (ZSL) \cite{palatucci2009zero} can be used to recognize unseen data by utilizing the relationship between seen and unseen data. In general, ZSL seeks to recognize unseen data by exploiting the correlation between seen and unseen data. This correlation is established using semantic knowledge, which can be obtained through human annotations \citep{lampert2009learning} or word-to-vector approaches \citep{mikolov2013efficient}. By using semantic descriptors, ZSL enables the transfer of information from seen to unseen domains. Generalized Zero-Shot Learning (GZSL) \cite{chao2016empirical} expands on ZSL by including additional seen classes in the target decision domain, and it has received increasing attention from researchers.

\begin{figure}[t]
	\centering
		\includegraphics[width=0.42\textwidth]{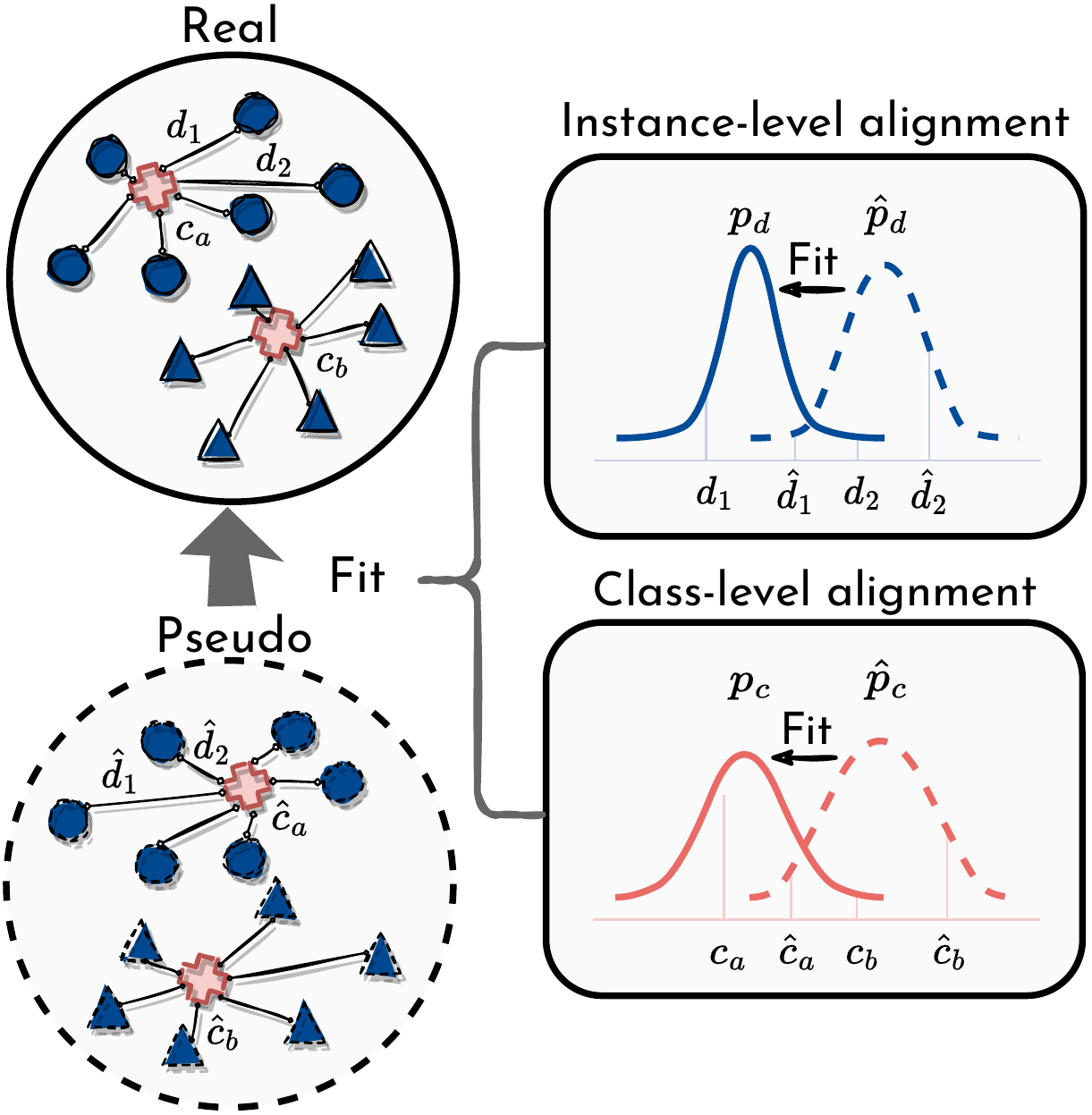}
% 	\vspace{-2ex}
	\caption{Illustration of two types of distributions learned by a generator: instance-level and class-level. $c$ represents the potential class center, while $d$ denotes an off-center position.}
    \label{fig:distribution}
	\vspace{-1ex}
\end{figure}

Recently, generative models have been used in mainstream GZSL research to supplement information on unseen classes. A central hypothesis of generation-based GZSL methods is that the generated class-level and instance-level unseen distribution should match the real unseen distribution (Fig. \ref{fig:distribution}). By generating pseudo-unseen instances, these methods enable classifier training to encompass unseen classes, resulting in a superior discrimination of unseen classes compared to their counterparts. Despite their success in enhancing GZSL performance, generation-based methods encounter various challenges in future extensions or developments. Firstly, the underlying reasons for the effectiveness of these approaches remain largely unexplored. Although certain literature suggests that improved discrimination \cite{wu2020self} or diversity \cite{liu2021near} of generated samples contributes to enhanced GZSL performance, no theoretical or empirical evidence supports these performance gains. Secondly, training a generative model entails additional computational and complexity. In most generation-based methods, the primary time complexity arises from training the generative model.

To address these challenges, we conduct both an empirical and a theoretical investigation to uncover, understand, and extend generation-based methods. We begin by analyzing the role of instance-level distribution and class-level distribution. In doing so, we replace the generator-learned instance-level distribution with the Gaussian distribution and conclude its substitutability in improving GZSL performance. (Sec. \ref{ea}). By decomposing the gradient of the cross-entropy loss, we further relate class- and instance-level distributions to unseen class discrimination and decision boundary formation (Sec. \ref{classifier}). Based on our analysis, we point out the core improvement direction for the generator-classifier framework. First, the key for the ZSL generator is attribute generalization, where we should focus on generalizing the attribute-conditioned image distribution learned from the seen data to unseen classes. Second, classifier learning is an independent task to learn from partially biased data. We summarize two principles for this task: mitigating the impact of pseudo samples on seen class boundaries during training and reducing the seen-unseen bias.

We finally propose a single baseline based on the idea of deconstruction. Our approach surpasses existing methods in performance, despite having lower complexity. Additionally, we replace the generative model with a one-to-one mapping network from attributes to the visual class centers. Our without-generator method retains most of the performance, which is a step towards simplifying the generator-classifier framework. Our main contributions include:
	\begin{itemize}
		\item We deconstruct the generator-classifier framework, using empirical and theoretical analysis to expose the core components of generator and classifier learning. 
\item We provide a guideline for optimizing the generator-classifier GZSL framework based on our deconstruction idea, which we use to derive a simple method.

\item Without a complicated framework design, the proposed method achieves SotAs on four popular ZSL benchmark datasets. Additionally, our method can also be transferred to other generative methods, even a single attribute-vision center mapping net, bringing us closer to a streamlined generator-classifier framework.
	\end{itemize}
\section{Related Work}
\noindent \textbf{Zero-Shot Learning (ZSL)}~\cite{lampert2009learning,farhadi2009describing} has been extensively studied in recent years, which requires knowledge transfer with the class-level edge information, \eg, human-defined attributes \cite{farhadi2009describing,parikh2011relative,akata2015evaluation} and word vectors \cite{mikolov2013efficient,mikolov2013distributed}. Traditional ZSL models \cite{akata2013label,frome2013devise} typically project the attribute and the visual feature to a common space. \citet{lampert2013attribute,frome2013devise,elhoseiny2013write} choose the attribute space as the common space. Some research afterward \cite{zhang2017learning,li2019rethinking,skorokhodov2020class} also embed attributes to visual space, or embed attributes and visual features to another space \cite{akata2015evaluation,zhang2015zero}. These methods achieve good performance in the classic ZSL setting but meet a \textbf{seen-unseen bias problem} (\ie, prediction results are biased towards seen classes) in \textbf{Generalized Zero-Shot Learning (GZSL)} \cite{chao2016empirical,xian2017zero} which emphasizes seen-unseen discrimination. 

Driven by the new technology in deep learning, some research enables deeper attribute-visual association with attribute attention \cite{zhu2019semantic,huynh2020fine,xu2020attribute,liu2021goal,wang2021dual}. Other methods introduce the out-of-distribution discrimination \cite{atzmon2019adaptive,min2020domain,chou2020adaptive}, which decomposes the GZSL task into seen-unseen discrimination and inter-seen (or -unseen) discrimination. The most successful methods in GZSL build on the recent advent of generative models \cite{goodfellow2014generative,kingma2013auto}, which have dominated recent ZSL research. The \textbf{generation-based methods} \cite{xian2018feature,xian2019f,chen2022zero} construct pseudo unseen samples to constrain the decision boundary, which form a better seen-unseen discrimination than their counterparts. 

A large amount of literature aims at improving the generation-based framework. \cite{xian2019f,shen2020invertible} focus their attention on new generative frameworks. \cite{verma2020meta} explores the training method. These methods do not make full use of the prior information in the ZSL setting but seek breakthroughs from other fields. \cite{narayan2020latent} design a recurrent structure that utilizes the intermediate layers of the visual-to-attribute mapping network for a second generation. \cite{han2020learning,han2021contrastive,chen2021free,chen2021semantics,kong2022compactness} propose to transform the visual feature into an attribute-dependent space, the pseudo unseen samples generated in which contain less seen class bias information. The above-mentioned methods usually adopt complex strategies, which trade large time consumption for performance. In this paper, we explore the nature of the generation-based framework, surpassing current SotAs without complex design.
\vspace{-0ex}\section{Generation-Based ZSL: A~Deconstruction }\label{deconstruction}
Assume there are two disjoint class label sets $\mathcal{Y}^{s}$ and $\mathcal{Y}^{u}$ ($\mathcal{Y}=\mathcal{Y}^{s}\cup{Y}^{u}$), ZSL aims at recognizing samples belong to $\mathcal{Y}^{u}$ while only having access to samples with the labels in $\mathcal{Y}^{s}$ during training. Denote $\mathcal{X} \subseteq \mathbb{R}^{d_x}$ and $\mathcal{A} \subseteq \mathbb{R}^{d_a}$ as visual space and attribute space, respectively, where $\mathbf{x} \in \mathcal{X}$ and $\mathbf{a} \in \mathcal{A}$ represent feature instances and their corresponding attributes (represented as column vectors) with dimensions $d_x$ and $d_a$. Given the training set $\mathcal{D}^{s}=\{ \mathbf{x},y,{\mathbf{a}}_{y} |\mathbf{x}\in \mathcal{X},y\in \mathcal{Y}^{s},{\mathbf{a}}_{y}\in \mathcal{A}\}$, the goal of ZSL is to learn a classifier towards the unseen classes: $f_{zsl}:\mathcal{X}\rightarrow\mathcal{Y}^{u}$. GZSL extends this to classify samples belonging to either seen or unseen classes, \ie, $f_{gzsl}:\mathcal{X}\rightarrow\mathcal{Y}$. We mainly discuss the challenges in the GZSL setting in this work.

In this paper, we focus on deconstructing the generator-classifier ZSL framework by understanding the behavior of the generator and the classifier. The framework involves training a conditional generator using visual-attribute pairs, followed by generating pseudo unseen samples using attributes from unseen classes. Finally, the ZSL or GZSL classifier is trained using the generated samples.

\vspace{-1ex}\subsection{Empirical Analysis of the Generator-Learned Instance-Level Distribution}\label{ea}
In generation-based methods, the generator is often relied upon to produce distributions for unseen classes. To better analyze the ZSL generator, we divide this distribution into two parts, as illustrated in Fig. \ref{fig:distribution}: the \textit{class-level distribution}, which determines how various unseen attributes are mapped to fit the real inter-class distribution in visual space, and the \textit{instance-level distribution}, which deals with how generated samples of the same unseen attribute fit the real intra-class distribution. As the class-level distribution is fundamental to inter-class discrimination, our analysis will concentrate on exploring the generator-fitted instance-level distribution. Specifically, we will compare it to other human-defined distributions based on fitness (against real distribution) and Zero-Shot performance.
\begin{table}[t]
\centering
\resizebox{0.47\textwidth}{!}{
\begin{tabular}{@{}llccccc@{}}
\toprule
Method                     & DIST & $\mathit{T_1}$ & $\mathit{A}^u$ & $\mathit{A}^s$ & $\mathit{H}$  & CMMD             \\ \toprule 
\multirow{4}{*}{f-CLSWGAN} & GEN         & 69.0           & 57.8           & 71.1           & 63.8          & \textbf{0.0337} \\
                           & SVG          & 68.2          &55.3 &71.7 &62.5        & 0.0341          \\
                           & LVG          & \textbf{69.7}  & 62.8  & 76.3  & \textbf{68.9} & 0.2523          \\
                           & SCG          & 69.5           & 62.8  & 68.5           & 65.5          & 0.0339          \\ \midrule
\multirow{4}{*}{CE-GZSL}   & GEN         & 69.8           & 63.5  & 77.5           & 69.8          & \textbf{0.0071} \\
                           & SVG          & 69.4           & 60.1           & 78.2           & 68.0          & 0.0099          \\
                           & LVG          & 66.0           & 47.9           & 72.7           & 57.7          & 0.2541          \\
                           & SCG          & \textbf{70.6}  & 63.2           & 78.9  & \textbf{70.2} & \textbf{0.0071} \\ \bottomrule
\end{tabular}
}
\caption{Zero-Shot performance and CMMD \wrt~different pseudo unseen distributions (DIST). \textbf{GEN:} Generated distribution; \textbf{SVG:} Small-variance Gaussian distribution; \textbf{LVG:} Large-variance Gaussian distribution; \textbf{SCG:} Statistical-covariance Gaussian distribution.}
\label{dist}
\vspace{-3ex}
\end{table}
\paragraph{Setup.} We conduct a comparison between the generator-fitted instance-level unseen distribution and three Gaussian distributions, which have independent small variance, independent large variance, and data-statistical covariance. Since typical zero-shot learning (ZSL) generators usually generate centralized distributions, we replace the instance-level distribution by shifting the centers of other distributions to the generated class centers. We then evaluate the Zero-Shot performance of these distributions and their discrepancy against real unseen distributions. The discrepancy is measured with Maximum Mean Discrepancy (MMD), which is a typical sample-based discrepancy measurement in research of domain adaptation \cite{long2015learning} and generative models \cite{tolstikhin2017wasserstein}. We calculate the MMD between the test unseen data and the experimental data for each class, and then take the average value to obtain the Centered MMD (CMMD) score:
\begin{equation}
   \resizebox{0.9\linewidth}{!}{$
	\begin{aligned}
	\mathrm{CMMD}=&\frac{1}{|\mathcal{Y}^u|}\sum_{c=1}^{|\mathcal{Y}^u|}\{\frac{1}{n_c\left(n_c-1\right)}\sum_{i,j=1,i\neq j}^{n_c}[\kappa\left(x_i^{c},{x}_j^c\right)\\
    +&\kappa\left(\widetilde{x}_i^{c},\widetilde{x}_j^c\right)]
    -\frac{2}{n_c^2}\sum_{i,j=1}^{n_c}\kappa\left(x_i^{c},\widetilde{x}_j^c\right)\},
	\end{aligned}
      \label{cmmd}
      $}
\end{equation}
where $x_i^c$ and $\widetilde{x}_i^{c}$ represent samples from class $c$ in the test unseen and pseudo unseen sets, respectively. $n_c$ denotes the sample number in class $c$, and $ \kappa\left(\cdot,\cdot \right)$ is generally an arbitrary positive-definite reproducing kernel function. Note that the test data involved here is only for measuring the distribution discrepancy and is not used in training.

\paragraph{Results.} We experiment with two classic generation-based methods, f-CLSWGAN~\cite{xian2018feature} and CE-GZSL~\cite{han2021contrastive}, on AWA2 dataset \cite{lampert2013attribute}. The results presented in Tab.~\ref{dist} led us to two main observations: (i) Gaussian distribution with statistical covariance produces similar results to the generated distribution in both methods; and (ii) the unrealistic unseen distribution negatively affects the performance of CE-GZSL but improves the performance of f-CLSWGAN. These observations prompted us to explore two questions: \textit{(i) Can we generate only the class center instead of using a complex generative model? (ii) How does the large-variance Gaussian distribution affect Zero-Shot performance?} We answer the first question experimentally in Sec. \ref{exp}, demonstrating that generating only class centers can still achieve reasonable Zero-Shot performance. To address the second question, we further investigate the role of pseudo unseen class samples in classifier training from a gradient perspective.

\begin{figure}[t]
	\centering
    \subfigure []
	{
		\includegraphics[width=0.219\textwidth]{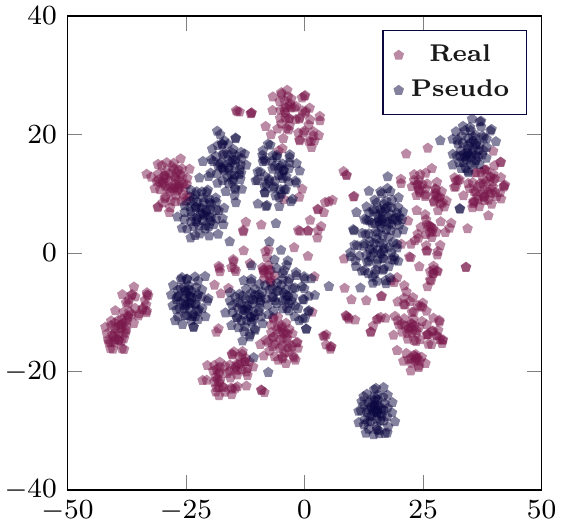}
	} 
    \subfigure []
	{
		\includegraphics[width=0.219\textwidth]{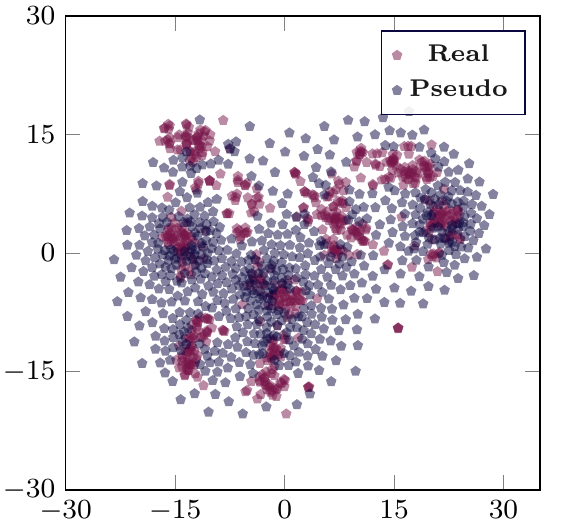}
	}
        \subfigure []
	{
		\includegraphics[width=0.219\textwidth]{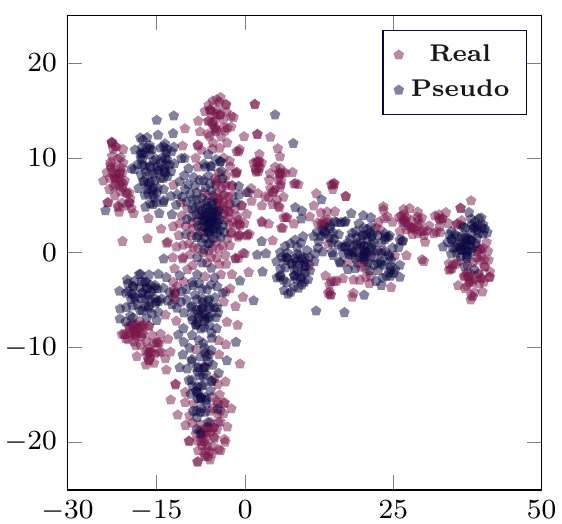}
	} 
    \subfigure []
	{
		\includegraphics[width=0.219\textwidth]{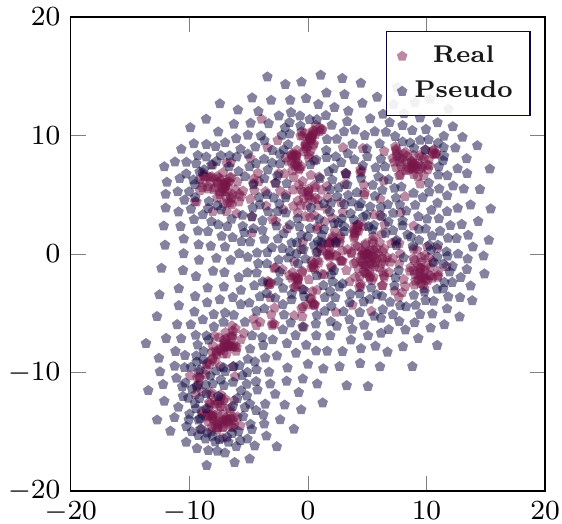}
	}
	\vspace{-2ex}
	\caption{t-SNE comparison across various pseudo unseen distributions. \textbf{(a)} Generated with f-CLSWGAN; \textbf{(b)} Large-variance Gaussian distribution moved to the class center generated with f-CLSWGAN; \textbf{(c)} Generated with CE-GZSL; \textbf{(d)} Large-variance Gaussian distribution moved to the class center generated with CE-GZSL.}
	\vspace{-3ex}
	\label{tsne}
\end{figure}

\vspace{-1ex}
\subsection{Impact of Pseudo Unseen Samples on Classifier Learning}\label{classifier}
% \vspace{-1ex}
We consider a linear classifier with weight parameters $\mathbf{W}\in\mathbb{R}^{|\mathcal{Y}|\times d_\mathbf{x}}$. With a slight abuse of notation, we subsequently use $(\mathbf{x},y)$ to denote both real and generated data. In the generation-based framework, the classifier is commonly trained using cross-entropy loss:
\begin{equation}
	\begin{split}
\mathcal{L}_{ce}&=\frac{1}{n}\sum_{c=1}^{|\mathcal{Y}|}\sum_{i=1}^{n_c}-\log p_y(\mathbf{x}_i)\\
p_y(\mathbf{x}_i)&=\frac{\exp(\langle\mathbf{W}_{y},\mathbf{x}_i\rangle/\tau)}{\sum_{c=1}^{|\mathcal{Y}|}\exp(\langle\mathbf{W}_c,\mathbf{x}_i\rangle/\tau)} .
	\end{split}
      \label{ce}
\end{equation}
Here, $\langle \cdot, \cdot \rangle$ denotes the dot product, $c$ is the index of the $c$-th row in $\mathbf{W}$, $n$ is the total number of samples, $n_c$ is the sample size in class $c$, and $\tau$ is the temperature parameter \cite{hinton2015distilling}.
\begin{theorem}
Gradients of $\mathcal{L}_{ce}$ can be decomposed into two components that indicate moving towards the class center and constraining the decision boundary, respectively:
\begin{equation}
\centering
% 	\resizebox{0.90\linewidth}{!}{$
		\displaystyle
	-\frac{\partial_{\mathcal{L}_{ce}}}{\partial_{\mathbf{W}_k}}=
		\left\{
		\begin{array}{lr}
		\frac{1}{n\tau}\sum_{i=1}^{n_k}\mathbf{x}_i\\
		-\frac{1}{n\tau}\sum_{c=1}^{|\mathcal{Y}|}\sum_{j=1}^{n_c} p_k(\mathbf{x}_j)\mathbf{x}_j\\
		\end{array},
		\right.
% 		$}
        \label{prop1}
	\end{equation}
where $p_k(\cdot)$ has an analogous definition to Eq. (\ref{ce}), and $\mathbf{W}_k$ represents the classifier weight of the $k$th class.
\label{prop}
\end{theorem}

The proofs of Proposition \ref{prop} is given in the appendix. According to Eq. (\ref{prop1}), \textit{the primary discriminant for unseen classes is determined by the fitness of the class-level distribution, while the instance-level pseudo unseen distribution controls the construction of decision boundaries.} Then we use Proposition \ref{prop} to analyze question \textit{(i)} of Sec. \ref{ea}. Specifically, we consider the seen-unseen bias problem where unseen class data is misidentified as seen class. A wider pseudo-unseen distribution promotes wider decision boundaries for unseen classes, which helps to mitigate the seen-unseen bias. As illustrated in Fig. \ref{tsne}, the large variance provides a wider pseudo unseen distribution for f-CLSWGAN that is still close to the real unseen distribution. In contrast, the feature distribution in CE-GZSL excessively deviates from the human-defined distribution as it uses a linear mapping on the original visual feature. From the perspective of decision boundaries, we can also understand the common strategy of sampling a large number of pseudo-unseen samples in classifier training \cite{xian2018feature,han2021contrastive}. An additional pseudo unseen datum $\mathbf{x}^u$ pulls class weight $\mathbf{W}^u$ towards the corresponding pseudo unseen distribution while pushing other class weights away, thus widening the unseen decision boundaries.

In conclusion, in Sec. \ref{deconstruction}, we deconstruct and summarize the essential aspects of the generator and classifier in generation-based methods. Next, we will provide explicit optimization guidelines founded on the above analysis.

\section{Generator-Classifier Learning under the Idea of Deconstruction}\label{gen-classi}
\label{att_genera}
\subsection{Learning Generator in Generalization View}
In Sec. \ref{ea}, we demonstrate that the generator-fitted instance-level unseen distribution is substitutable in Zero-Shot recognition. Therefore, we suggest focusing on optimizing the class-level distribution, which serves as the core to guide the gradient (Eq. (\ref{prop1})). To improve the class-level distribution, we provide insights from a generalization perspective. In typical supervised classification tasks, generalization refers to the learned conditional probability $q(y|\mathbf{x})$ from the empirical distribution $p(\mathbf{x},y)$ fitting the test set. Inspired by this, we propose attribute generalization as the key to ZSL generator:

\begin{theorem}[Key to ZSL generator]
\label{theo_prop2}
Attribute generalization in Zero-Shot generation is the conditional probability $p_g(\mathbf{x}|\mathbf{a})$ modeled on $p_r^s(\mathbf{x},\mathbf{a}|\mathbf{a}\in \mathcal{A}^s)$ fitting $p_r^u(\mathbf{x},\mathbf{a}|\mathbf{a}\in \mathcal{A}^u)$, where $p_r^s$ and $p_r^u$ are the real seen and unseen distributions, respectively.
\end{theorem}

By converting a distributional learning problem into a generalization problem, we can handle it directly with existing tools. Drawing from established research on generalization problems in supervised classification tasks, we examine various existing overfitting suppression strategies such as L2 regularization, the Fast Gradient Method \cite{goodfellow2014explaining} (an adversarial training method), and attribute augmentation. These techniques lead to improvements in the original generator's Zero-Shot performance, as well as the CMMD (Eq. (\ref{cmmd})) against real unseen data. For further information and additional experiments on attribute generalization, please see Sec. \ref{sec:additional} in the appendix.

\subsection{Learning Classifier with Partly Biased Data}\label{principle}
Due to the absence of unseen class data in ZSL setting, the generated unseen class data are bound to deviate from the real distribution, as shown in Fig. \ref{tsne}. Consequently, the main challenge in classifier learning is to capture the true decision boundary using partially biased data. However, data bias is unpredictable, and thus, it is essential for the classifier to adapt more toward the deterministic (\ie, real seen) distribution and reduce the adverse effects of biased (\ie, pseudo unseen class) distributions. Building upon the discussion in Sec. \ref{classifier}, we propose two principles for classifier design: (i) mitigating the impact of pseudo unseen samples on decision boundaries between seen classes during training, and (ii) reducing the seen-unseen bias.
\subsection{A Simple Method over the Guidelines}
We propose a simple method for verifying the validity of the above guidelines for generator-classifier learning. Our approach employs the widely-used \cite{gulrajani2017improved} as the generative model, which consists of a generator $\mathit{G}$ and a discriminator $\mathit{D}$ and is optimized by the following objective:
\begin{equation}
	\begin{split}
	\mathcal{L}&=\mathbb{E}_{\mathbf{x}\sim p_r}[\mathit{D}(\mathbf{x},\mathbf{a})]-\mathbb{E}_{\widetilde{\mathbf{x}}}[\mathit{D}(\widetilde{\mathbf{x}},\mathbf{a})]\\
    -\lambda_0\mathbb{E}_{\hat{\mathbf{x}}\sim p_{\hat{\mathbf{x}}}}&[(\nabla_{\hat{x}} \Vert\mathit{D}(\hat{\mathbf{x}},\mathbf{a})\Vert_2)^2-1] ,~\widetilde{\mathbf{x}}=\mathit{G}(\mathbf{z}_0,\mathbf{a}),
	\end{split}
      \label{eq1}
\end{equation}
where $p_r$ denotes the real distribution of $\mathbf{x}$, $\mathbf{z}_0\in\mathcal{N}(\mathbf{0},\mathbf{I})$, $\hat{\mathbf{x}}=\alpha \mathbf{x}+(1-\alpha)\widetilde{\mathbf{x}}$ with $\alpha \sim\mathit{U}(0,1) $ is for calculating the gradient penalty and $\lambda_0$ is a hyper-parameter.

We augment the attribute with Gaussian noise to enhance the attribute generalization (Proposition \ref{theo_prop2}), \ie,
\begin{equation}
	\begin{split}
\mathit{G}(\mathbf{z}_0,\mathbf{a})\rightarrow\mathit{G}(\mathbf{z}_0,\mathbf{a}+\mathbf{z}_1),
	\end{split}
      \label{eq3}
\end{equation}
where $\mathbf{z}_1\in\mathcal{N}(\mathbf{0},\sigma\mathbf{I})$, and $\sigma$ decides the standard deviation of the augmenting distribution. The reason for attribute augmentation is detailed in Sec. \ref{sec:additional} of the appendix.

During the classifier training phase, we follow \textit{principle (i)} (Sec. \ref{principle}) and begin by representing the unseen class corresponding terms in loss function as an increment (on the cross-entropy with seen class only), \ie,
\begin{equation}
	\begin{split}
&\mathcal{L}_{ce}=\frac{1}{n}[\sum_{c^s=1}^{|\mathcal{Y}^s|}\sum_i^{n_{c^s}}-\log\frac{p_y(\mathbf{x}_i)}{\hat{p}^s(\mathbf{x}_i)+\lambda_1\hat{p}^u(\mathbf{x}_i)}\\
+\lambda&_2\sum_{c^u=1}^{|\mathcal{Y}^u|}\sum_j^{n_{c^u}}-\log p_y(\mathbf{x}_j)],~\hat{p}^{\,\bigcdot}(\mathbf{x}_i)=\sum_{c=1}^{|\mathcal{Y}^{\,\bigcdot}|}p_{c}(\mathbf{x}_i),
	\end{split}
      \label{eq6}
\end{equation}
where $p_c(\cdot)$ is defined in Eq. (\ref{ce}), (\ref{prop1}). We introduce two parameters, $\lambda_1$ and $\lambda2$, to weight the generalized incremental forms. When $\lambda_1$ and $\lambda_2$ are set to zero, it indicates that the added pseudo unseen samples do not affect the seen class decision boundaries, and \textit{principle (i)} can be achieved by selecting small values for $\lambda_1$ and $\lambda_2$.
\begin{figure}[t]
	\centering
		\includegraphics[width=0.32\textwidth]{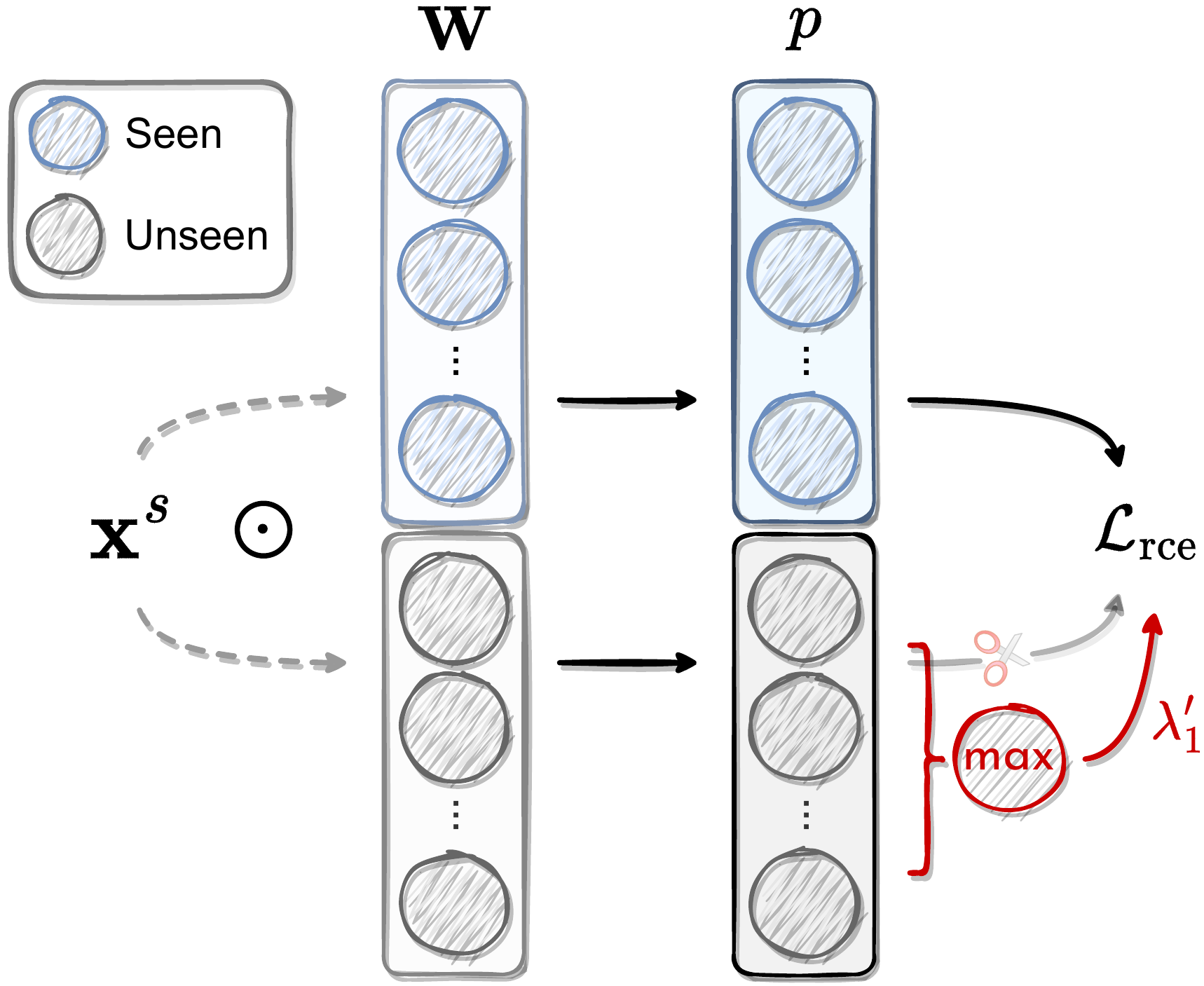}
% 	\vspace{-2ex}
	\caption{Illustration of the revised cross-entropy loss (Eq.~(\ref{rce})), where ${\odot}$ denotes the dot product. Per seen class sample, only the unseen class weight that gives it the largest activation is involved in the calculation. The calculation of the seen class weights remains unchanged.}
	\vspace{-3ex}
	\label{fig_rce}
\end{figure}

Then we express the gradient of $\mathcal{L}_{ce}$ with respect to the weights of an unseen class, $\mathbf{W}_u$, as
\begin{equation}
\centering
% 	\resizebox{0.90\linewidth}{!}{$
% 		\displaystyle
        	\begin{aligned}
	-\frac{\partial_{\mathcal{L}_{ce}}}{\partial_{\mathbf{W}_{u}}}&=\frac{\lambda_2}{n\tau}(\sum_{i=1}^{n_{u}}\mathbf{x}_i-\sum_{c^u=1}^{|\mathcal{Y}^u|}\sum_{j=1}^{n_{c^u}} p_u(\mathbf{x}_j)\mathbf{x}_j)\\
   - &\frac{1}{n\tau}\sum_{c^s=1}^{|\mathcal{Y}^s|}\sum_{k=1}^{n_{c^s}} \frac{\lambda_{1}p_u(\mathbf{x}_k)}{\hat{p}^s(\mathbf{x}_k)+\lambda_1\hat{p}^u(\mathbf{x}_k)}\mathbf{x}_k .
	\end{aligned}
        \label{eq:gradient}
% 		$}
	\end{equation}
Here, a small $\lambda_1$ makes the seen data have little effect on the decision boundaries of unseen classes, while $\lambda_2$ determines the extent to which the loss function focuses on inter-unseen-class decision boundaries. This provides a direction to mitigate the seen-unseen bias, \ie, the \textit{principle (ii)}.

In summary, selecting a small value for $\lambda_1$ and an appropriate value for $\lambda_2$ aligns with the two guiding principles for classifier design. As $\lambda_2$ has the same optimization direction as the generation number of pseudo-unseen samples, we remove it by fixing it to 1. We assign different values of $\lambda_1$ to each unseen class based on their optimization difficulty. Empirically, we only set non-zero values for the hardest class, and only if it exceeds the true class score, as illustrated in Fig. \ref{fig_rce}. The revised cross-entropy formula is presented as:

\begin{equation}
\resizebox{0.90\linewidth}{!}{$
	\begin{aligned}
&\mathcal{L}_{rce}=\frac{1}{n}\sum_{c^s=1}^{|\mathcal{Y}^s|}\sum_i^{n_{c^s}}-\log\frac{p_y(\mathbf{x}_i)}{\hat{p}^s(\mathbf{x}_i)+\lambda_1'p_m^u(\mathbf{x}_i)}\\
+\sum_{c^u=1}^{|\mathcal{Y}^u|}&\sum_j^{n_{c^u}}-\log p_y(\mathbf{x}_j),~p_m^u(\mathbf{x}_i)=\max\left\{p_c(\mathbf{x}_i)| c \in \mathcal{Y}^{u}\right\},
	\end{aligned}
      \label{rce}
      $}
\end{equation}
where $\lambda_1'=\lambda_1\mathds{1}[p_m(\mathbf{x}_i>p_y(\mathbf{x}_i)]$, and $\mathds{1}[\cdot]$ is the indicator function. The classifier trained with an appropriate value of $\lambda_1$ exhibits stronger inter-seen class discriminability and smaller seen-unseen bias, as demonstrated in Fig. \ref{para} (c), (d). Finally, we constrain the classifier weights with the attributes using a mapping network $\mathit{M}(\cdot)$, \ie,
\begin{equation}
	\begin{split}
\mathbf{W}_c:=\mathit{M}(\mathbf{a}_c),c\in\mathcal{Y}^s\cup\mathcal{Y}^u,
	\end{split}
      \label{eq:mapping net}
\end{equation}
which replaces the weights in Eq. (\ref{ce}). We also normalize the elements before feeding them into the dot product, which is a common strategy in ZSL. After training, a datum $\mathbf{x}$ is classified as the class with the attribute exhibiting the greatest similarity to it, \ie,
\begin{equation}
	\begin{split}
\hat{y}=\arg\max_{c}\langle\frac{\mathit{M}(\mathbf{a}_c)}{||\mathit{M}(\mathbf{a}_c)||_2},\frac{\mathbf{x}}{||\mathbf{x}||_2}\rangle,
	\end{split}
      \label{eq10}
\end{equation}
where $||\cdot||_2$ denotes the $l_2$ norm.

\begin{table*}[t]
	\centering
	\resizebox{0.9\textwidth}{!}{
		\begin{tabular}{@{}cllccc|ccc|ccc|ccc@{}}
        \toprule
			&\multirow{2}{*}{Method} & \multirow{2}{*}{Source}                                                     & \multicolumn{3}{c}{AWA2}                                                       & \multicolumn{3}{c}{CUB}                                                        & \multicolumn{3}{c}{SUN}                                                        & \multicolumn{3}{c}{APY}                      \\
			&       &                     & \multicolumn{1}{|c}{$\mathit{A}^u$}                           &$\mathit{A}^s$                    & $\mathit{H}$                         & $\mathit{A}^u$                        &$\mathit{A}^s$                      & $\mathit{H}$                      & $\mathit{A}^u$                     & $\mathit{A}^s$                    & $\mathit{H}$                         & $\mathit{A}^u$            & $\mathit{A}^s$            &$\mathit{H}$             \\
            \noalign{\smallskip}
            \hline
            \noalign{\smallskip}
\multicolumn{1}{c|}{\multirow{9}{*}{$\dagger$}}&       Chou et al.  &\multicolumn{1}{|l|}{ICLR \citeyear{chou2020adaptive}}       &65.1 &78.9 &71.3  &41.4 &49.7 &45.2    &29.9 &40.2 &34.3   &35.1 &65.5 &45.7\\
         \multicolumn{1}{c|}{}&   SDGZSL                  & \multicolumn{1}{|l|}{ICCV \citeyear{chen2021semantics}}     & 64.6 & 73.6 & 68.8                     & 59.9                     & 66.4                     & 63.0                     & 48.2	&36.1	&41.3                       & 38.0                     & 57.4                     & 45.7                     \\

    \multicolumn{1}{c|}{}&        GCM-CF  &\multicolumn{1}{|l|}{CVPR \citeyear{yue2021counterfactual} }       &60.4 &75.1 &67.0  &61.0 &59.7 &60.3    &47.9 &37.8 &{42.2}   &37.1 &56.8 &44.9\\
		\multicolumn{1}{c|}{}&	CE-GZSL                & \multicolumn{1}{|l|}{CVPR \citeyear{han2021contrastive}}         & 63.1          & 78.6          & 70.0          & 63.9    & 66.8          & 65.3          & 48.8         & 38.6          &43.1          & -             & -             & -             \\
                  \multicolumn{1}{c|}{}&  SE-GZSL       &\multicolumn{1}{|l|}{AAAI \citeyear{kim2022semantic}}       &59.9 &80.7  &68.8& 53.1 &60.3& 56.4&45.8 &40.7 &43.1&- &-&-\\
                  \multicolumn{1}{c|}{}&  ICCE       &\multicolumn{1}{|l|}{CVPR \citeyear{kong2022compactness}}        &65.3 &82.3&\underline{72.8} &67.3 &65.5 &66.4 &-&-&-&45.2&46.3&45.7\\
                  		\multicolumn{1}{c|}{}&	ZLA                & \multicolumn{1}{|l|}{IJCAI \citeyear{chen2022zero}}         &65.4	&82.2	&\underline{72.8}          & 73.0	&64.8	&\underline{68.7}          &50.1	&38.0	&\underline{43.2}         & 40.2	&53.8	&46.0           \\
                  
            \noalign{\smallskip}
         \cline{2-15}
              \noalign{\smallskip}
		\multicolumn{1}{c|}{}&	\textbf{DGZ}      & \multicolumn{1}{|l|}{\multirow{2}{*}{\textbf{Proposed}}}        & 67.4   & 81.0        & \textbf{73.6}   & 70.1& 68.3          & \textbf{69.2} & 48.6          & 39.4          &  \textbf{43.5}    & 37.7    & 64.9          &  \textbf{47.7}    \\ 
        		\multicolumn{1}{c|}{}&	\textbf{DGZ w/o GM}      & \multicolumn{1}{|l|}{}        & 65.9   &78.2          & 71.5  & 71.4&64.8     &68.0     &49.9      & 37.6        &   42.8 &38.0    &63.5         &\underline{47.6}     \\ 
           \noalign{\smallskip}
         \hline
         \hline
           \noalign{\smallskip}
        		\multicolumn{1}{c|}{\multirow{9}{*}{$\ddagger$}}&	TF-VAEGAN*              & \multicolumn{1}{|l|}{ECCV \citeyear{narayan2020latent}}                & 55.5 &83.6 &66.7          &63.8 &79.3&70.7          & 41.8 &51.9& \underline{46.3}          & -             & -             & -             \\
                \multicolumn{1}{c|}{}&       Chou et al.*  &\multicolumn{1}{|l|}{ICLR \citeyear{chou2020adaptive}}       &69.0 &86.5&\underline{76.8}  &69.2& 76.4 &72.6    &50.5& 43.1& \textbf{46.5}    &36.2& 58.6& 44.8\\
                \multicolumn{1}{c|}{}		&GEM-ZSL   &\multicolumn{1}{|l|}{ CVPR \citeyear{liu2021goal}  }                   	
				&64.8& 77.5& 70.6
                			&64.8 &77.1& 70.4 
				&38.1& 35.7 &36.9 
						&-&-&-\\
                                  \multicolumn{1}{c|}{}&   SDGZSL*                  & \multicolumn{1}{|l|}{ICCV \citeyear{chen2021semantics}}     & 69.6 &78.2 &73.7                     & 73.0 &77.5 &75.1                     & 51.1	&40.2	&45.0                      & 39.1&60.7 &47.5                  \\
                                  \multicolumn{1}{c|}{}		&DPPN      & \multicolumn{1}{|l|}{NeurIPS \citeyear{wang2021dual} }       			
				& 63.1 &86.8& 73.1
				& 70.2 &77.1 &73.5
				&   47.9 &35.8 &41.0
				&40.0 &61.2 &48.4\\
            	\multicolumn{1}{c|}{}&		TransZero                  &\multicolumn{1}{|l|}{AAAI \citeyear{Chen2022TransZero} }                &  61.3 &82.3 &70.2   & 69.3 &68.3 &68.8    &52.6&33.4 &40.8             & -       & -          & -          \\
                                                             \multicolumn{1}{c|}{}&  MSDN       &\multicolumn{1}{|l|}{CVPR \citeyear{chen2022msdn}}        &62.0& 74.5 &67.7 &68.7 &67.5& 68.1 &52.2 &34.2 &41.3&-&-&-\\
                                                                         \noalign{\smallskip}
         \cline{2-15}
              \noalign{\smallskip}
                                                                     		\multicolumn{1}{c|}{}&	\textbf{DGZ*}      & \multicolumn{1}{|l|}{\multirow{2}{*}{\textbf{Proposed}}}                  & 71.7 &83.7&\textbf{77.2}    & 76.9& 77.7         & \underline{77.3} &49.4         & 43.5          &  \underline{46.3}   &  37.1    & 79.3         &  \textbf{50.5}    \\ 
                                                                                                                                                 		\multicolumn{1}{c|}{}&	\textbf{DGZ* w/o GM}      & \multicolumn{1}{|l|}{}                  &67.2 &85.7 &75.4     & 77.4&78.0  &\textbf{77.7}     &50.4         & 39.8         &  44.5   &  38.5   &  67.4  &  \underline{49.0} \\ 
                           
            \bottomrule
            \end{tabular}
            }
                	\caption{GZSL performance comparison with state of the arts. $\dagger$ denotes generative methods based on the common image feature proposed in \citet{xian2017zero}. $\ddagger$ denotes allowing fine-tuning the feature extraction backbone, and * represents generative methods based on features extracted from the fine-tuned backbone. $\mathit{A}^u$ and $\mathit{A}^s$ are per-class accuracy scores (\%) on seen and unseen test sets. $\mathit{H}$ is their harmonic mean. The best results are shown in bold, with second place underlined.}
     \label{tab2}
	\vspace{-3ex}
\end{table*}
\section{Experiments}\label{exp}
\noindent {\bf Benchmark Datasets.} We conduct GZSL experiments on four public ZSL datasets. Animals with Attributes 2 (AWA2) \cite{lampert2013attribute} contains 50 animal species and 85 attribute annotations, accounting 37,322 samples. Attribute Pascal and Yahoo (APY) \cite{farhadi2009describing} includes 32 classes of 15,339 samples and 64 attributes. Caltech-UCSD Birds-200-2011 (CUB) \cite{wah2011caltech} consists of 11,788 samples with 200 bird species, annotated by 312 attributes. SUN Attribute (SUN) \cite{patterson2012sun} carries 14,340 images from 717 different scenario-style with 102 attributes. We split the data into seen and unseen classes according to the common benchmark procedure in \citet{xian2017zero}.

\noindent {\bf Representation.} Most experiments are performed with the 2048-dimensional visual features extracted from the pre-trained ResNet101 \cite{he2016deep}, following \citet{xian2017zero}. We also compare the GZSL performance on the fine-tuned data that we take from \citet{chen2021semantics}. For class representations (\ie, attributes), we adopt the artificial attribute annotations that come with the datasets for AWA2, APY, and SUN, and employ the 1024-dimensional character-based CNN-RNN features \cite{reed2016learning} generated from textual descriptions for CUB.

\noindent {\bf Evaluation Metric.} We calculate the average per-class top-1 accuracy among the unseen and seen classes respectively, denoted as $\mathit{A}^u$ and $\mathit{A}^s$, then their harmonic mean $\mathit{H}$ is employed as the measurement of GZSL. The classic ZSL is evaluated with per-class averaged top-1 accuracy on unseen classes \cite{xian2017zero}.

\noindent{\textbf{Implementation Details.}} The method proposed in Sec.~\ref{gen-classi} consists of three modules implemented with multi-layer perceptrons. The Generator $\mathit G$ carries two hidden layers with 4096 and 2048 dimensions. The Discriminator $\mathit{D}$ contains one 4096-D hidden layer, and the mapping net $\mathit{M}$ includes a 1024-D hidden layer. All the hidden layers are activated by Leaky-ReLU. We follow \citet{xian2018feature} to set other hyper-parameters of WGAN-GP. In addition, we put 512 for the (mini) batch size and adopt Adam \cite{kingma2014adam} as the optimizer with a learning rate of $1.0 \times 10^{-4}$.

\subsection{Comparison with SotAs\label{sota}}
We evaluate the proposed method by comparing its GZSL results with the current SotAs, as shown in Tab. \ref{tab2}. Notably, our results on common image features outperform SotAs in all four datasets. Moreover, our fine-tuned feature results ranked first on three datasets and second only to \citet{chou2020adaptive} on SUN dataset. It is important to highlight that our approach is simple and does not require complex designs. Yet, it outperforms other complex approaches such as \citet{chou2020adaptive}, which uses the out-of-distribution discrimination method, and \citet{han2021contrastive,kong2022compactness}, which rely on instance discrimination, both leading to significant time consumption.

We also report the results without a generative model. The pseudo unseen distribution is constructed as mixed Gaussian distribution with the covariance as the statistics of the training set. A one-to-one mapping net (from attributes to visual class centers) estimates its mean (detailed in the appendix). In this baseline, our method still achieves comparable performance with current SotAs. It demonstrates the plug-in capability of the proposed classifier learning strategy, even in the case of no generator. It is also an attempt to simplify the generator-classifier framework.

\begin{table}[t]
		\resizebox{0.473\textwidth}{!}{
			\begin{tabular}{llcccccc}
\toprule
\multirow{2}{*}{} & \multirow{2}{*}{Ablation} & \multicolumn{3}{c}{AWA2}                       & \multicolumn{3}{c}{CUB}                        \\
\multicolumn{2}{c}{}                                             & $\mathit{A}^u$ & $\mathit{A}^s$ & $\mathit{H}$ & $\mathit{A}^u$ & $\mathit{A}^s$ & $\mathit{H}$ \\
\midrule
   (i)          & w/o ATA                   & 66.4           & 77.2           & 71.4         & 72.2          & 66.1           & 69.0         \\
  (ii)        & w/o CR                   & 39.8           & 89.4           & 55.1         & 58.3           & 70.9          & 64.0         \\
(iii)                                 & w/o M                    & 64.0           & 79.4           & 70.9         & 70.7           & 57.8           & 63.6 \\  (iv)                               &w/o CR\&M                    & 34.7&90.0&50.0         & 44.7           & 70.2           & 54.7          \\ 
(v)                               &DIST $\rightarrow$ SCG                  & 67.5&78.0&72.4         & 68.3           & 67.7           & 68.0          \\ 
(vi)                               &DIST $\rightarrow$ GC+SCG                  & 65.9&78.2&71.5         & 71.4           & 64.8           &68.0          \\ 
\midrule
                                  & \textbf{Full Model}                &67.4             &81.0                &\textbf{73.6}              & 70.1           & 68.3           & \textbf{69.2}         \\ 
          \bottomrule
\end{tabular}
	}
            	\caption{Ablation study results on AWA2 and CUB. The baselines are constructed by ablating some key modules. \textbf{ATA:} Attribute augmentation; \textbf{CR:} Classifier revision; \textbf{M:} Mapping net. \textbf{SCG:} Statistical-covariance Gaussian distribution. \textbf{GC:} Direct generating the class center.}
                     \label{tab:ablation}
                     \vspace{-3ex}
\end{table}
\subsection{Ablation Study}\label{sec:ablation}
\noindent{\textbf{Baselines.}} To validate the effect of each component, we conduct an ablation study on AWA2 and CUB, with the following baselines: \textbf{(i)} Setting $\sigma$ to $0$. \textbf{(ii)} Training the classifier with vanilla cross-entropy. \textbf{(iii)} Removing the mapping net (Eq. (\ref{eq:mapping net})). \textbf{(iv)} Combination of (ii) and (iii). \textbf{(v)} Replacing the WGAN-generated distribution with the statistical-covariance Gaussian distribution (same to Sec. \ref{ea}). \textbf{(vi)} On the basis of (v), directly estimating the mean of the distribution by mapping from the attributes (same to Sec. \ref{sota}).

\noindent \textbf{Results.} Tab. \ref{tab:ablation} depicts the results of this experiment. {\textbf{Baseline (i)}} shows that the fewer effects of attribute augmentation on the fine-grained dataset CUB than on the coarse-grained dataset AWA2. This is mainly due to the fine-grained dataset's inherently smaller domain shift problem, causing less gain from a targeted approach. Meanwhile, for the same reason, classifier revision plays a bigger role for AWA2 than for CUB (\textbf{baseline (ii), (iv)}). \textbf{Baseline (iii), (iv)} reflect the importance of the mapping net, which establishes implicit semantic connections between classifier weights. Overall, due to its intractability, attribute generalization enhancement brings fewer performance gains than classifier revision. \textbf{Baseline (v) and (vi)} compare the ways to obtain the mean of Gaussian distribution. Baseline (v) averages the WGAN-generated samples for the mean of each class, which yields better performance than directly mapping attributes to the class mean (baseline (vi)). This is probably because the instance-level modeling extracts more distribution information and better generalizes to unseen class attributes. More details and analysis are Provided in the appendix.
\begin{figure}[t]
	\centering
   
        \subfigure []
	{
		\includegraphics[width=0.22\textwidth]{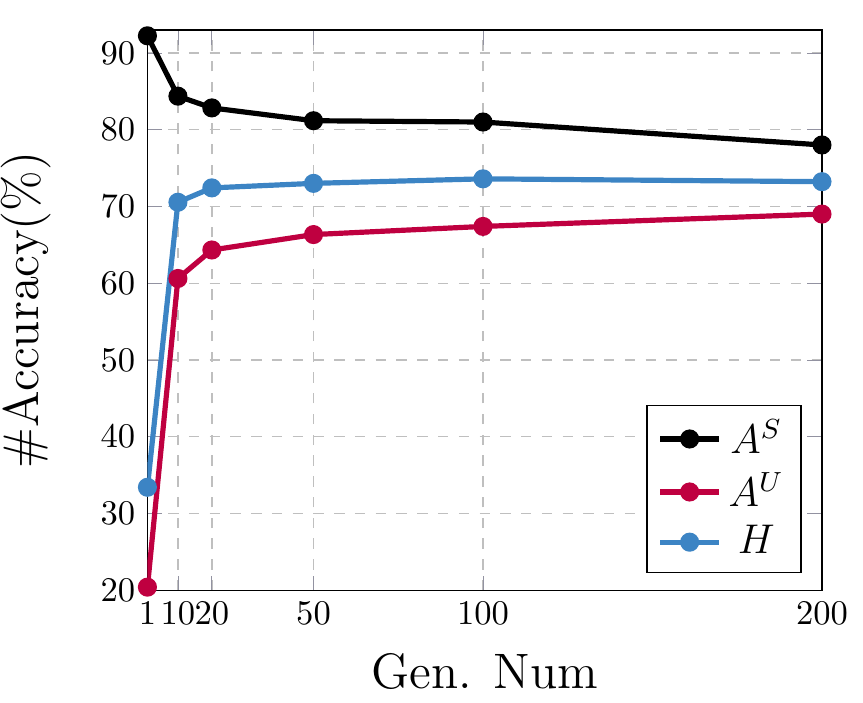}
	}
     \subfigure []
	{
		\includegraphics[width=0.22\textwidth]{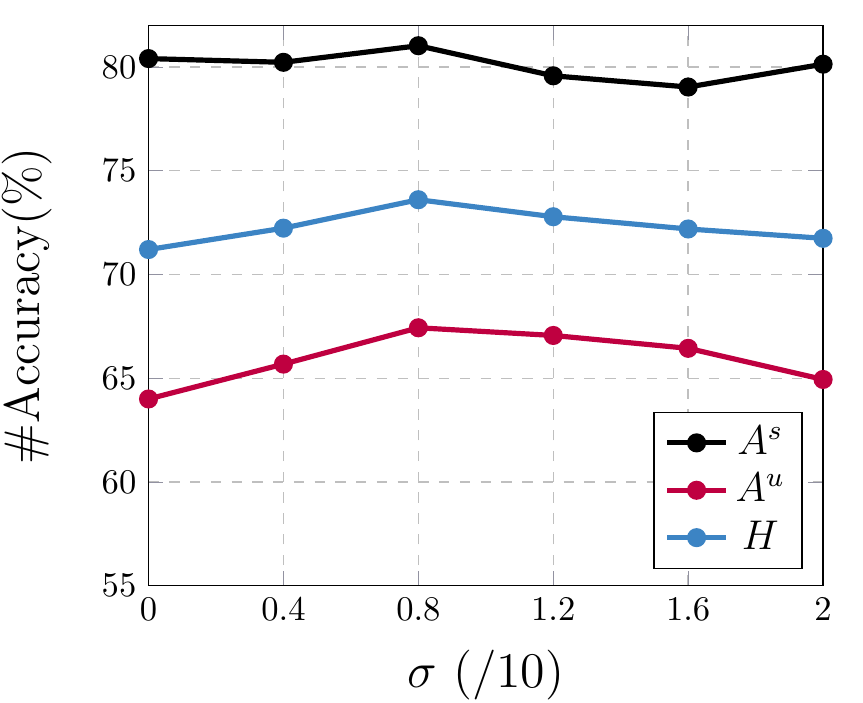}
	}
    \subfigure []
	{
		\includegraphics[width=0.2225\textwidth]{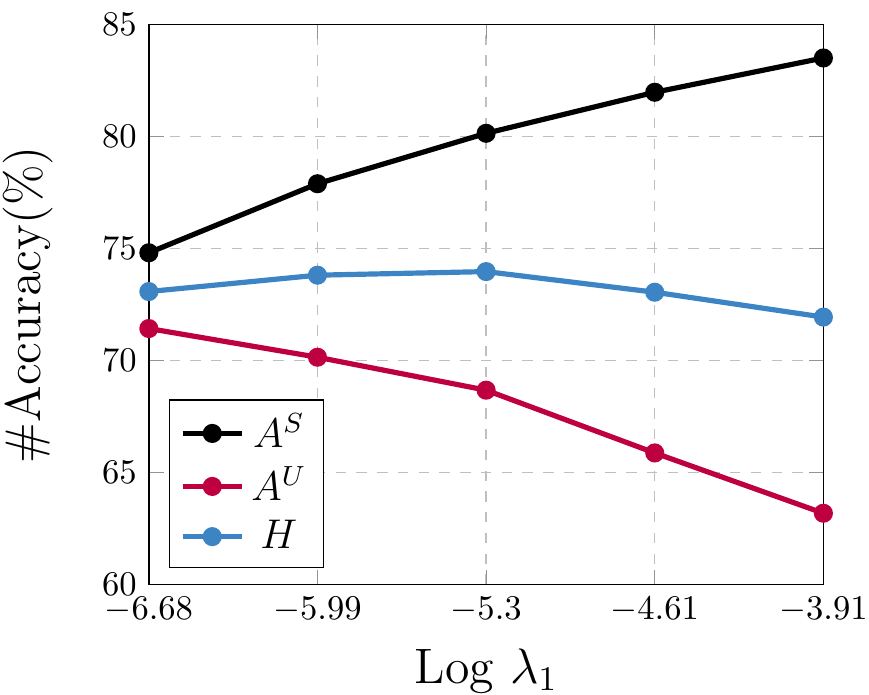}
	}
    	\subfigure []
	{
		\includegraphics[width=0.2225\textwidth]{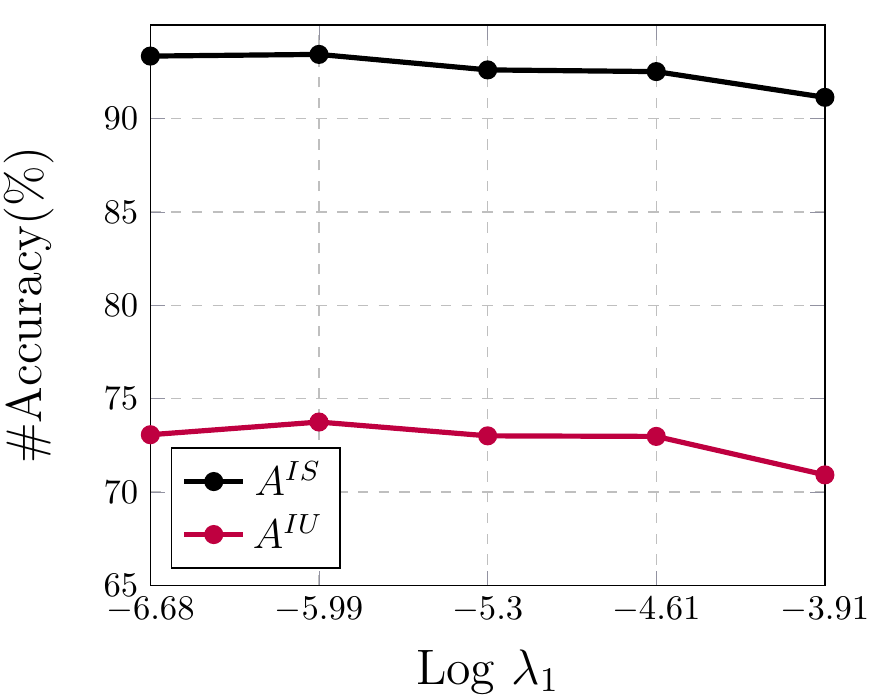}
	}
	\vspace{-2ex}
	\caption{\textbf{(a), (b), (c)} GZSL performance \wrt~the generation number per unseen class, $\sigma$, and $\lambda_1$. \textbf{(d)} Intra-discriminability of seen and unseen classes \wrt~$\lambda_1$, where $\mathit{A}^{is}$ and $\mathit{A}^{iu}$ represent the intra- seen or unseen classes accuracy. The experiments are conducted on AWA2 dataset.}
	\vspace{-3ex}
	\label{para}
\end{figure}
\subsection{Hyper-parameters}
The final objective involves four main hyperparameters: $\sigma$, $\tau$, $\lambda_1$, and the generated number per unseen class. We set $\tau$ to 0.04, following \citet{skorokhodov2020class,chen2022zero}. We then analyze the influence of the other three parameters empirically. As shown in Fig. \ref{para} (b), $\mathit{A}^u$ and $\mathit{H}$ have the same trend when $\sigma$ varies, whose curves rise first and then fall as $\sigma$ becomes larger. A big $\sigma$ leads to performance degradation because a large variance of noise intuitively makes the attribute input of the generator lose inter-class discriminability. A small $\lambda_1$ mitigates the seen-unseen bias in Fig. \ref{para} (c). Moreover, a suitable generated number creates the best performance, as shown in Fig. \ref{para} (a), and the number is much smaller than the existing generation-based methods (100 \textit{vs.} 2400 in \cite{han2021contrastive} and 4600 in \cite{chen2021free}). This demonstrates the joint effect of the number of generations and $\lambda_1$ as we stated in Sec. \ref{gen-classi}. We also report the effect of $\lambda_1$ on the intra-seen class discriminability in Fig. \ref{para} (d), showing a downward trend when $\lambda_1$ increases within a certain range. We empirically generate $50$ samples per unseen class in CUB, SUN, and APY, and 100 for AWA2 in all experiments. We put $\lambda_1$ to 4, 0.8, 0.04, and 0.005 for the above datasets. $\sigma$ is set to $0.08$ on all datasets.

\subsection{Discriminability on Unseen Classes}
As shown in Tab. \ref{zsl}, we analyze the discriminability of the trained GZSL classifier among unseen classes, quantified by ZSL accuracy. Despite not being specifically designed for the ZSL setting, our model still achieves comparable results to SotA ZSL methods. This is primarily due to improvements in attribute generalization ability and the intrinsic semantic association of classifier weights carried from attribute mapping.
\begin{table}[t]
\centering
	\resizebox{0.4\textwidth}{!}{
% 		{\spaceskip=0.15em\relax
				\begin{tabular}{lcccc}
					\toprule
					Method                                         & AWA2          & CUB           & SUN           & APY           \\ 
                    \midrule
					TCN \shortcite{jiang2019transferable}                & 71.2          & 59.5          & 61.5          & 38.9          \\
					TF-VAEGAN \shortcite{narayan2020latent}             & 72.2          & 64.9          &\textbf{66.0}    & -             \\
                    Chou et al. \shortcite{chou2020adaptive} &73.8   &57.2  &63.3  &41.0\\
					IPN \shortcite{liu2021isometric}                 & \textbf{74.4 }         & 59.6          & -             & 42.3          \\
                    CE-GZSL \shortcite{han2021contrastive}          & 70.4          & 77.5  & 63.3          & -             \\
                     SDGZSL \shortcite{chen2021semantics}  &72.1 &75.5 &- &45.4\\
                   \midrule
					\textbf{DGZ}                   & 74.0& \textbf{80.1} & 65.4          & \textbf{46.6} \\
                    \bottomrule
				\end{tabular}
                	}
                		\caption{Discriminability on unseen classes, evaluated by ZSL performance (\%) (compared with SotAs). \textbf{Note} that our classifier is trained towards the GZSL setting.}
        \label{zsl}
			\vspace{-3ex}
% 		}
\end{table}
\vspace{-1ex}
\section{Conclusion}
In this paper, we deconstruct the generator-classifier Zero-Shot Learning framework. We begin by decomposing the unseen class distribution learned by the generator into class- and instance-level distribution. Then we empirically analyze the learning center of the generator and the role of these two distributions in classifier learning. Specifically, we emphasize attribute generalization in generator training and regard classifier training as an independent task to learn from partially biased data. Based on these points, we propose a simple method that outperforms current SotAs in performance without a complex design, demonstrating the effectiveness of the proposed guideline. Additionally, we evaluate the transferability of the proposed method and find that it can achieve SotA even when replacing the generative model with a class center mapping net. We acknowledge that our analysis is primarily empirical and lacks mathematical discussion. We will explore the generation-based framework more thoroughly from a theoretical standpoint and continue to simplify it in future work.
\vspace{-1ex}
\section*{Acknowledgements}
This work was partly supported by the National Natural Science Foundation of China (NSFC) under Grant Nos. 61872187, 62077023, and 62072246, partly by the Natural Science Foundation of Jiangsu Province under Grant No. BK20201306, and partly by the ``111 Program" under Grant No. B13022.

\bibliography{aaai23}
\clearpage

\appendix
\renewcommand\thefigure{A.\arabic{figure}} 
\renewcommand\theequation{A.\arabic{equation}} 
\renewcommand\thetable{A.\arabic{table}} 

\section*{Appedix}
\section{Experimental Details}
\subsection{Empirical Analysis in Sec \ref{ea}}
We generate 4000 and 2400 pseudo unseen samples for experiments in f-CLSWGAN and CE-GZSL, respectively, which corresponds to the number reported in the published paper. To assess the Gaussian distribution, we begin by computing the center of the pseudo unseen class using the generated samples. We then sample from the Gaussian distribution and move the result to the pseudo centers.

We employ the Inverse Multiquadratic (IM) kernel \cite{ardizzone2018analyzing,shen2020invertible}: $ \kappa\left(\mathbf{x}, \mathbf{x}^{\prime}\right)=2 d_{x} /\left(2 d_{x}+\left\|\mathbf{x}-\mathbf{x}^{\prime}\right\|^{2}\right)$ to calculate MMD. We randomly generated samples for each unseen class with the same number as the test set. MMD is a metric that assesses the similarity between two distributions, and a value closer to zero indicates a higher similarity between the two.

\subsection{Regarding the One-to-One Mapping Net}
In Sec. \ref{sota} and Sec. \ref{sec:ablation}, we conduct experiments with a one-to-one mapping network from attributes to visual centers. To achieve this, we minimize the Mean Squared Error (MSE) loss between the mapped visual centers and the visual instances, \ie,
\begin{equation}
	\begin{split}
&\mathcal{L}_{mse}=\frac{1}{n}[\sum_{c=1}^{|\mathcal{Y}^s|}\sum_i^{n_{c}}||\mathit{G}(\mathbf{a}_{c})-\mathbf{x}_i||_2.
	\end{split}
      \label{eq:oto}
\end{equation}
The hyper-parameter used to train the one-to-one mapping network is the same as the one used to train WGAN.

\begin{figure}[t]
	\centering
    \subfigure []
	{
		\includegraphics[width=0.22\textwidth]{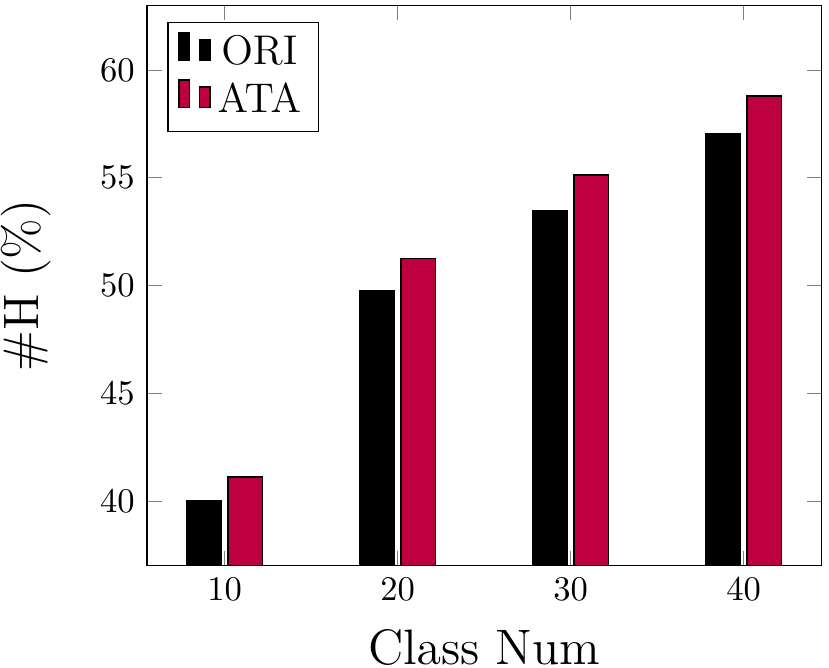}
	}
    	\subfigure []
	{
		\includegraphics[width=0.22\textwidth]{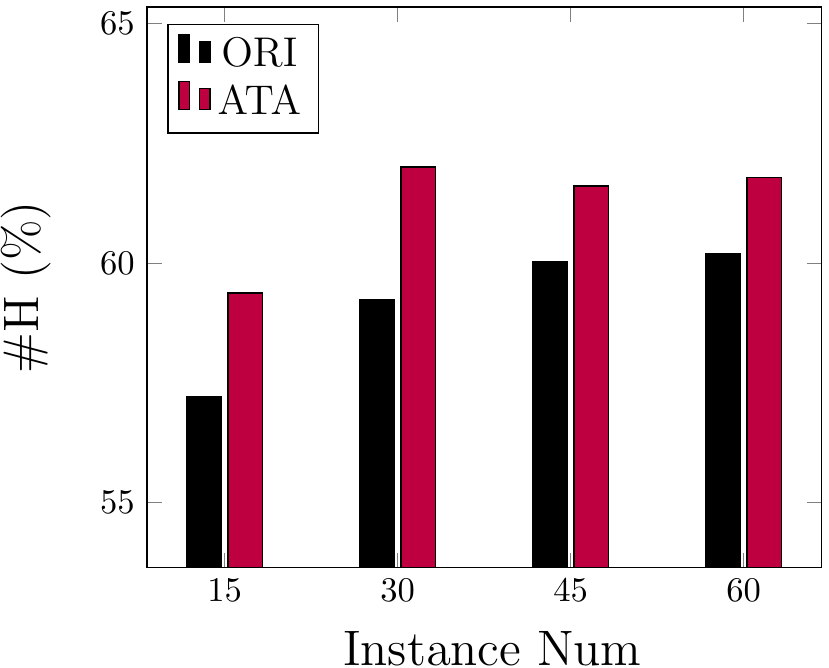}
	}
    \subfigure []
	{
		\includegraphics[width=0.224\textwidth]{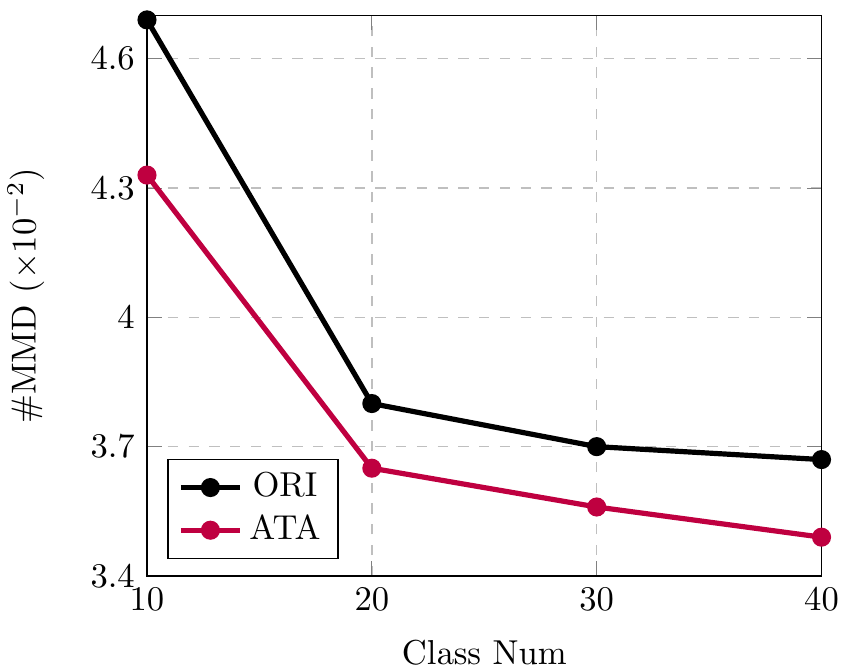}
	}
    \subfigure []
	{
		\includegraphics[width=0.224\textwidth]{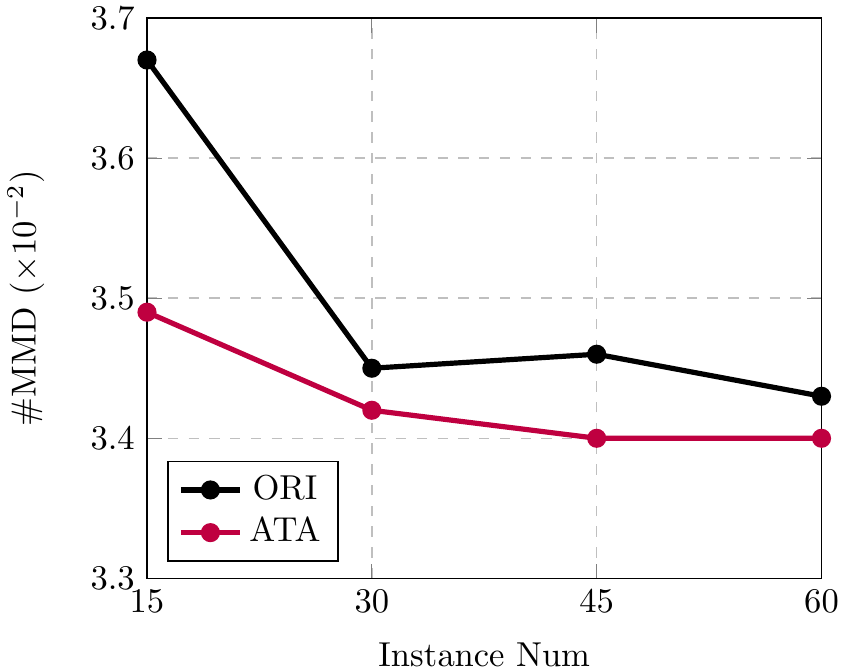}
	}
	\vspace{-1ex}
	\caption{Empirical analysis on the generalization bound of the original WGAN (averaged by ten random experiments on AWA2). \textbf{(a), (c):} $H$ score and MMD value \wrt the amount of classes. \textbf{(b), (d):} $H$ score and MMD value \wrt the amount of visual instance involved in training. {\bf ORI} and \textbf{SA} denote the original and the attribute augmentation results, respectively.}
	\vspace{-1ex}
	\label{gen_bound}
\end{figure}
\section{Additional Experiments}\label{sec:additional}
\subsection{Analysis on Attribute Generalization}\label{sec_ea}
We additional analyze the bound of attribute generalization in WGAN, and the effect of typical overfitting suppression strategies on attribute generalization. The generalization of WGAN is evaluated with GZSL performance (the higher the better) and CMMD value (the lower the better).
\paragraph{Empirical Analysis on Generalization Bound.}
We performe a series of comparative experiments to investigate the attribute generalization bound in Zero-Shot generation. In supervised classification tasks, the classifier's generalization ability is largely determined by the diversity of the generalizing objects, such as images in image classification. Similarly, we examine the diversity of generalizing objects (attributes) in the Zero-Shot generation task. Our study consists of three parts. Firstly, \textit{(I) we maintaine a constant number of visual features (to reduce the impact of visual feature diversity) and randomly sample different numbers of classes (where one class is associated with one attribute) to compare the generalization results.} However, since changes in the number of classes could affect the diversity of visual features, we proceed to conduct the other two experiments. Secondly, \textit{(ii) we fix the number of classes and increase the number of visual features.} Thirdly, \textit{(iii) we keep the number of visual features constant and augment the input generator's attributes with Gaussian noise (similar to image augmentation in other vision tasks), \ie,} 

\begin{equation}
\nonumber
\resizebox{0.90\linewidth}{!}{$
	\begin{aligned}
\mathit{G}(\mathbf{z}_0,\mathbf{a})\rightarrow\mathit{G}(\mathbf{z}_0,\mathbf{a}+\mathbf{z}_1),\mathbf{z}_0\in\mathcal{N}(\mathbf{0},\mathbf{I}),\mathbf{z}_1\in\mathcal{N}(\mathbf{0},\sigma\mathbf{I}) ,
	\end{aligned}
      \label{eq_aug}
      $}
\end{equation}
where $\sigma$ decides the standard deviation of the sampled distribution. The experiments are conducted on the original WGAN-GP \cite{gulrajani2017improved}, and the results are plotted in Fig. \ref{gen_bound}.

As shown in Fig. \ref{gen_bound} (a), (c), improved GZSL performance and a smaller MMD value can be obtained when the amount of classes involved in training grows. This is in line with our expectations because a similar conclusion that increasing the number of generalization objects can produce generalization gains has been drawn in classification tasks. In the meanwhile, the increase in the class number brings richer visual features. However, Fig. \ref{gen_bound} (b), (d) show that the increased number of visual features quickly saturates the generalization gain, which indicates its limited benefit on generalization. Surprisingly, attribute augmentation yields considerable positive results in all cases, demonstrating that merely increasing attribute diversity enhances attribute generalization, shown as the red histograms and lines in Fig. \ref{gen_bound}. This phenomenon, together with the observation of the above experiments, shows that attribute diversity is an important factor of attribute generalization.

In experiment (i), we randomly select 10, 20, 30, and 40 classes from the seen class set. The number of the sampled visual instances is fixed at 600 and divided evenly among the sampled classes. In experiment (ii), we set the number of sampled classes to 40, and for the visual instances, each class is sampled 15, 30, 45, and 60 times. In experiment (iii), we augment the attribute input of the generator with Gaussian noise ($\sigma=0.08$). We generate 4000 examples for each unseen class after training the generator, then combine the generated samples with the real seen samples to train the classifier.
\begin{table}[t]
    \centering
\begin{tabular}{lccccc}
\hline\noalign{\smallskip}
Method &T1& $\mathit{A}^u$    & $\mathit{A}^s$    & $\mathit{H}$&CMMD   \\ 
\noalign{\smallskip}
\hline
\noalign{\smallskip}
ORI    &68.0 & 57.2 & 70.4 & 63.1 &0.0345\\
L2 Norm &71.1& 59.3&\textbf{72.4}&65.2 &0.0337\\
FGM     &\textbf{73.4}& 60.8&71.2&65.6 &\textbf{0.0333}\\
ATA    &72.7  & \textbf{61.7}&70.7&\textbf{65.9}&0.0335\\ 
\hline
\end{tabular}
     	\caption{Effects of overfitting suppression methods on the generator, measured on AWA2. {\bf ORI}: the original WGAN. {\bf L2 Norm}: L2 regularization. {\bf FGM}: Fast Gradient Method. {\bf ATA}: Attribute augmentation.}
\label{tab_over}
\end{table}
\paragraph{Testing Existing Overfitting Suppression Strategies.}
The definition of attribute generalization in Proposition \ref{theo_prop2} is analogous to feature generalization in supervised classification. In this study, we consider their same gain pattern as analyzed in Sec. \ref{sec_ea} and investigate the performance of different overfitting suppression methods on attribute generalization. Tab. \ref{tab_over} shows that L2 regularization, Fast Gradient Method \cite{goodfellow2014explaining} (an adversarial training method), and attribute augmentation all improve upon the original WGAN. It is noteworthy that basic attribute augmentation produces results comparable to the more complex FGM strategy.

To implement L2 regularization, we introduce an additional loss function that minimizes the F norm of all the weight matrices in the generator. We set the coefficient to 0.001 and add the loss function to the generator's existing loss. For attribute augmentation, we add a noise sampled from a Gaussian distribution with a standard deviation of 0.08 to the attribute input of the generator. To apply FGM, we calculate the gradient from the generator loss to the attribute input and use this gradient to update the attribute with a learning rate of 0.08. The updated attribute is then fed into the generator to calculate the loss for updating its parameters. Finally, after training the generator, we generate 4000 samples per unseen class for classifier training.

\begin{figure}[t]
	\centering
     	\subfigure []
	{
		\includegraphics[width=0.215\textwidth]{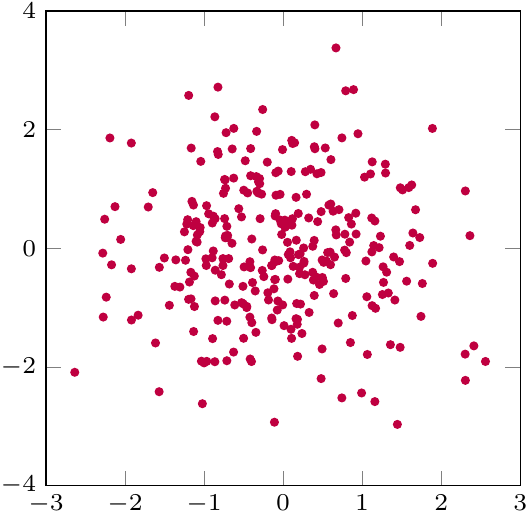}
	}
    \subfigure []
    	{
		\includegraphics[width=0.22\textwidth]{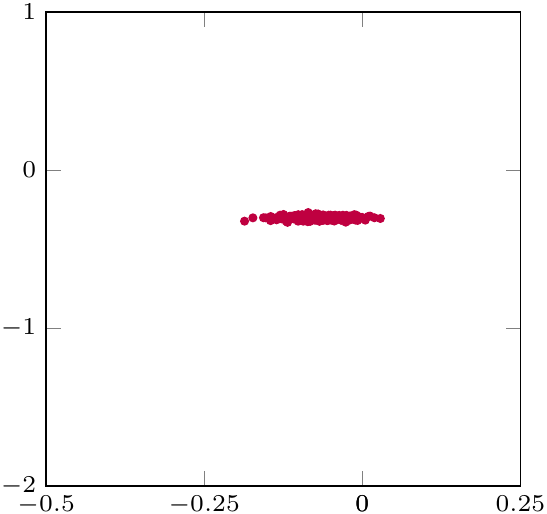}
        }
    	\subfigure []
	{
		\includegraphics[width=0.22\textwidth]{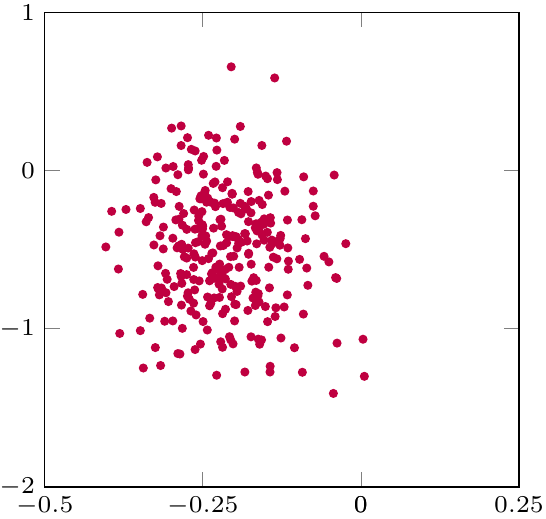}
	}
       	\subfigure []
	{
		\includegraphics[width=0.22\textwidth]{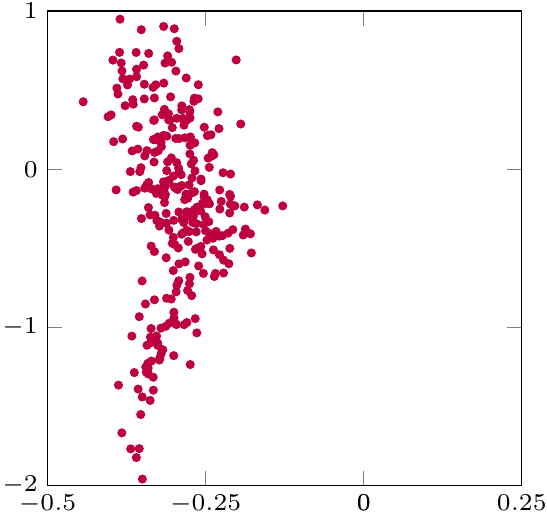}
	}

	\vspace{-1ex}
	\caption{Illustration of the toy experiment. \textbf{(a)} 2-D groundtruth sampled from the Unit Gaussian distribution. \textbf{(b) (c)} WGAN-generated samples without or with attribute augmentation during training. \textbf{(d)} Attribute augmentation with large variance.}
	\vspace{-1ex}
	\label{fig_toy}
\end{figure}
\subsection{Toy Experiment: Why Attribute Augmentation Effective?}
\label{sec_toy}
Despite the effectiveness of the attribute augmentation (ATA) approach, we further conduct a toy experiment to illustrate its mechanism.
\begin{table*}[t]
\centering
% \vspace{-1.8ex}
\resizebox{0.7\textwidth}{!}{
\begin{tabular}{@{}lcccccccccccc@{}}
\toprule
\multirow{2}{*}{Method} & \multicolumn{3}{c}{AWA2}                       & \multicolumn{3}{c}{CUB}                        & \multicolumn{3}{c}{SUN}                        & \multicolumn{3}{c}{APY}                        \\
                        & $\mathit{A}^u$ & $\mathit{A}^s$ & $\mathit{H}$ & $\mathit{A}^u$ & $\mathit{A}^s$ & $\mathit{H}$ & $\mathit{A}^u$ & $\mathit{A}^s$ & $\mathit{H}$ & $\mathit{A}^u$ & $\mathit{A}^s$ & $\mathit{H}$ \\ 
                        \midrule
VAE                     & 54.8           & 69.4           & 61.2         & 61.8           & 59.7           & 60.7         & 50.5           & 32.6           & 39.6         & 30.5           & 60.5           & 40.6         \\
VAE+DGZ                & 64.2           & 81.6           & \textbf{71.9}         & 68.4           & 69.0           & \textbf{68.7}         & 49.9           & 38.0           & \textbf{43.2}         & 37.7           & 65.0           & \textbf{47.7}   
\\ \bottomrule
\end{tabular}
}
\caption{Evaluation of plugging our method into CVAE.}
\label{vae}
\vspace{-1ex}
\end{table*}

\paragraph{Setup.} We consider a 2-D dataset sampled from Unit Gaussian distribution, \ie, $\mathcal{D}^{toy}=\{\mathbf{x}_i\}, \mathbf{x}_i\in\mathcal{N}(\mathbf{0},\mathbf{I})$. We define the coordinates on the 2-D plane as the attributes, an attribute vector defines a Gaussian distribution centered on it. We only attach to 1 attribute in this case, \ie, $\mathbf{a}=[0,0]$. The toy data are plotted in Fig. \ref{fig_toy} (a). Then we sample a noise $\mathbf{z}$ from an additional prior distribution $\mathcal{N}(\mathbf{0},\mathbf{I})$ to concatenate with $\mathbf{a}$ and train WGAN. We set $\sigma=0.04$ for Fig. \ref{fig_toy} (c) and $\sigma=1$ for Fig. \ref{fig_toy} (d). We sample 2000 points from the Unit Gaussian distribution for the dataset construction. After training, we generate 300 samples for observing the result.

\paragraph{Toy Experiment Analysis.} 
After training, we sample various $\mathbf{a}$ values from $\mathcal{N}(\mathbf{0},\mathbf{I})$ while fixing $\mathbf{z}$ to $[0,0]$ to show the generation results. In our setup, the attribute corresponds to a position on a 2-D plane. However, the synthesized results depicted in Fig. \ref{fig_toy} (b) without ATA are shrinks to a narrow range. This is because the model, which was trained with a single attribute, cannot capture the full range of attribute information. Conversely, Fig. \ref{fig_toy} (c) displays a Gaussian distribution with the ATA technique. In our dataset, each sample point is considered a distinct attribute. When augmenting the attribute in the input of the generator with a small variance Gaussian noise, the samples generated by the augmented attributes are constrained to the groundtruth space via optimization. As with generative models, where additional prior noise is regularly mapped into the feature space, the small variance of attributes automatically corresponds to locations in the visual space. An anomalistic distribution is observed in Fig. \ref{fig_toy} (d), which is trained by ATA with large variance noises. Intuitively, the feature generated with large variance attributes creates a high probability of being outside the groundtruth domain, which also breaks the rules of the setup.

\paragraph{Real Scenario Analysis.} 
Assuming that all important visual properties are described by attributes, ideally, an attribute vector can uniquely identify a location in the visual space. However, in the ZSL setting, we can only obtain the collective description of each class. The generative model is only able to learn the positional differences of large attribute gaps (\ie, attributes of different categories) in the visual space and is unable to capture the response to small attribute changes. By augmenting attributes with small variance noises and constraining the generated visual features within their corresponding categories through optimization, small attribute changes find the direction of change in the visual space. As a result, the generator learns additional implicit attribute-visual relations, which act as an inductive bias to improve its generalization ability on unseen attributes.

\subsection{Testing Our Method on VAE}
We plug our method into CVAE to test its transferring ability. We test it with the common visual feature proposed in \citet{xian2017zero} with the same hyper-parameter settings as in WGAN. As shown in Tab. \ref{vae}, our method yields significant performance gains over the vanilla setting (\ie, training vanilla cross-entropy with re-sampling pseudo unseen samples), again demonstrating its transferability.
\begin{table}
% \hspace{1ex}
    \centering
\begin{tabular}{lcccccc}
\hline\noalign{\smallskip}
\multirow{2}{*}{Method} & \multicolumn{3}{c}{AWA2}     & \multicolumn{3}{c}{CUB}                        \\
          & $\mathit{A}^u$ & $\mathit{A}^s$ & $\mathit{H}$ & $\mathit{A}^u$ & $\mathit{A}^s$ & $\mathit{H}$ \\
\noalign{\smallskip}
\hline
\noalign{\smallskip}
CN    &69.4&77.2&73.1 & 68.1&64.8&66.4\\
ours &67.4 &81.0 &\textbf{73.6} &70.1 &68.3 &\textbf{69.2}\\
\hline
\end{tabular}
     	\caption{We replace the weight mapping net with the advanced method \cite{skorokhodov2020class}, denoted as \textbf{CN}, and compare the results.}
        \vspace{-1ex}
\label{tab_cn}
\end{table}
\subsection{Replacing the Weight Mapping Net}
We note that some advanced classifier-weight mapping frameworks, known as prototype learning, have been developed as non-generative approaches. To demonstrate that we are not taking advantage of these methods, we replace the weight mapping net (implemented with a 3-layer MLP in our method) with an advanced method \cite{skorokhodov2020class} and compare the results. As shown in Tab. \ref{tab_cn}, the advanced mapping technique can not achieve a higher result than a simple net. This might be because such a strategy is specially designed for non-generative settings, leaving a gap in its application to generative settings (more specific reasons are orthogonal to our research).
\begin{figure*}
\centering
\subfigure[]{
\includegraphics[width=.25\textwidth]{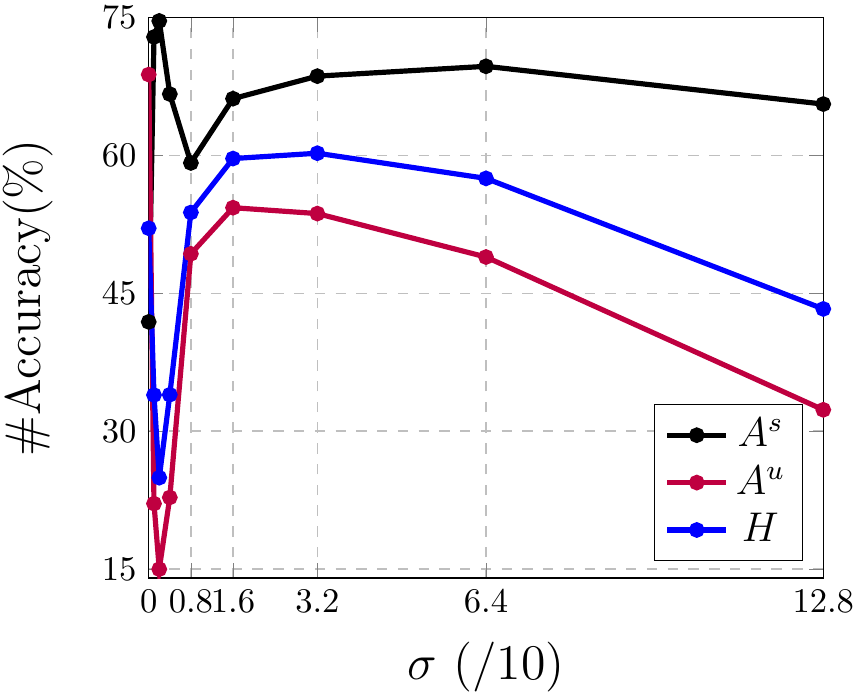}
}
\subfigure[]{
\includegraphics[width=.325\textwidth]{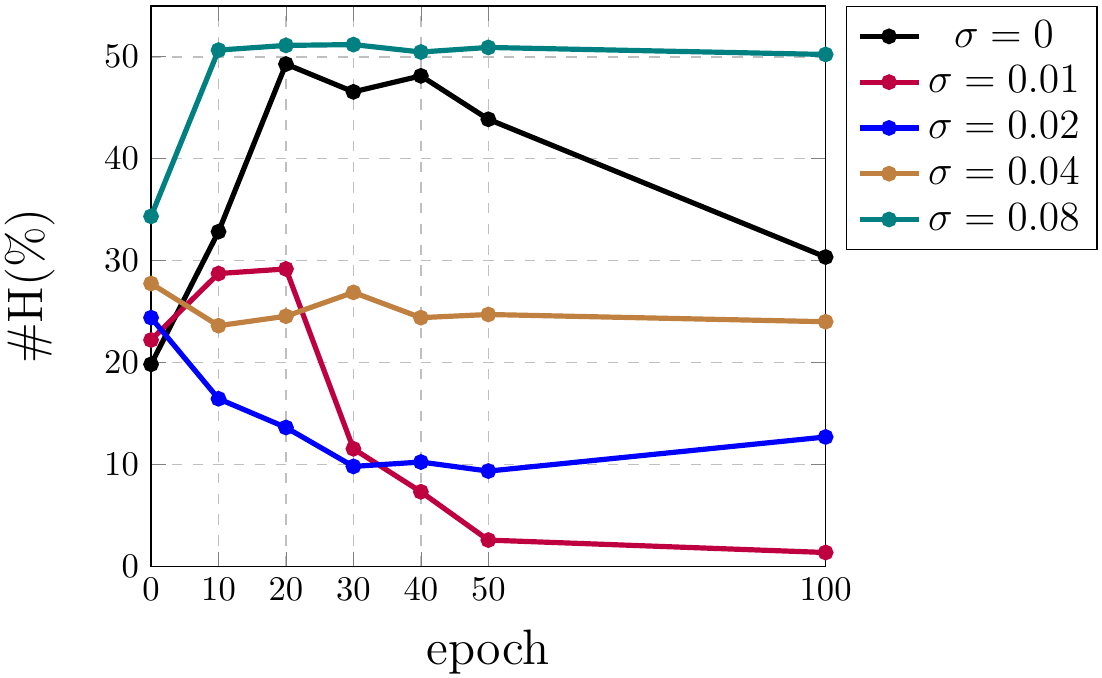}
}
\subfigure[]{
\includegraphics[width=.257\textwidth]{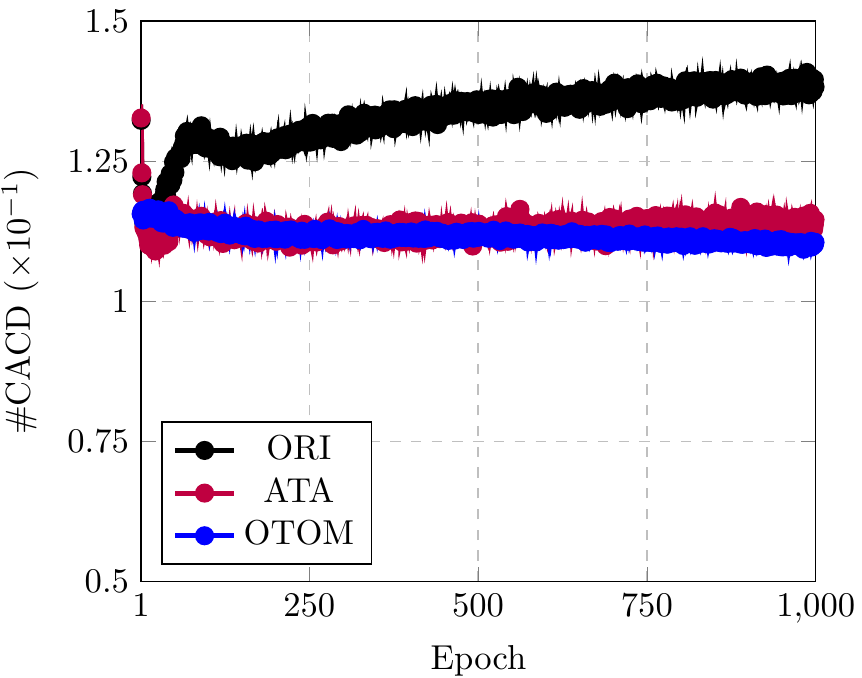}
}
\caption{\textbf{(a), (b):} We remove the additional prior input in the generator (\ie, $\mathbf{z}_0$), and compare the GZSL performance for attribute augmentation with varied standard variance ($\sigma$). \textbf{(c):} Class averaged center distance (CACD) comparison on ORI (original WGAN), ATA (WGAN+attribute augmentation), and OTOM (one-to-one mapping net).}
\label{replace prior}
\end{figure*}
\subsection{Experiment on Prior Distribution of Generators}
Traditional conditional generative models treat the attribute as a class generative controller and include additional prior distributions (\ie, $\mathbf{z}_0$) for implementing the generating process. With attribute augmentation, the additional prior distribution appears to be less essential. Hence, we discard $\mathbf{z}_0$ and evaluate the GZSL performance with various $\sigma$ on AWA2. As shown in Fig. \ref{replace prior} (a), the GZSL performance initially lowers, then rises, and finally declines again when $\sigma$ increases from 0. We attribute the initial GZSL performance to the fact that the generative model approximates a single mapping to the class center when $\sigma=0$. When the $\sigma$ grows slightly, the Gaussian distribution of its composition is insufficient to sustain the WGAN prior distribution, resulting in GAN's frequent training difficulties (e.g., pattern collapse and unstable training). It can also explain the unstable GZSL performance as the training progresses when $\sigma$ is small, as shown in Fig. \ref{replace prior} (b). GZSL performance suffers when $\sigma$ is large because excessive variance noise confuses the conditional discriminating ability of attributes. Overall, only employing the augmented attributes as the prior has a performance penalty compared to creating a new prior $\mathbf{z}_0$.
\subsection{Training Dynamics \wrt Attribute Generalization}
In this section, we employ the class averaged center distance (CACD) to indicate attribute generalization and evaluate the training dynamics of attribute generalization on the vanilla WGAN, WGAN+attribute augmentation, and the attribute-visual-center one-to-one mapping net. CACD is calculated by
\begin{equation}
\centering
% 	\resizebox{0.90\linewidth}{!}{$
		\displaystyle
	CADA=\sum_{k=i}^{\mathcal{Y}^u}\sqrt{||C_k^g-C_k^r||_2},
% 		$}
        \label{cada}
	\end{equation}
where $C_k^r$ denotes the real unseen class center, which is obtained by averaging the samples in the corresponding class. $C_k^g$ is the generated class center, which is obtained by averaging the generated samples in WGAN and is directly generated in the one-to-one mapping net. As shown in Fig. \ref{replace prior} (c), vanilla WGAN produces a high value of CACD, which reflects its weak attribute generalization ability. WGAN with attribute augmentation and the one-to-one mapping net have a similar trend in CACD variation, which demonstrates the effectiveness of attribute generalization. We evaluate CACD on AWA2.

\begin{figure*}[t]
\begin{lstlisting}[numbersep=5pt, frame=lines, framesep=2mm]
    def rev_cross_entropy(input,label,weight,tau,lambda_1,sclasses,uclasses):
        logits = input @ weight.t() / tau
        idx_s = torch.eq(label.reshape(-1, 1), sclasses).sum(1).nonzero().squeeze(1)
        mask = torch.ones_like(logits)
        mask[idx_s] = mask[idx_s].scatter (1, uclasses.repeat(len(idx_s), 1), 0)
        _, index = torch.max(logits, dim=1, keepdim=True)
        mask_ = torch.scatter(mask, 1, index, lambda_1)
        mask = (1 - mask) * mask_ + mask
        posi = logits[torch.arange(logits.size(0)).long(),label].view(-1,1)
        logits = logits - posi
        logits = torch.exp(logits)
        logits = mask * logits
        loss = (torch.log(logits.sum(1))).mean()
        return loss
\end{lstlisting}     
\caption{Implementation of the proposed revised cross-entropy in PyTorch.}
\label{code}
\end{figure*}

\section{Proof of Proposition \ref{prop}}
First, we recall the Proposition:
\begin{theorem}
Gradients of $\mathcal{L}_{ce}$ can be decomposed into two components that indicate moving towards the class center and constraining the decision boundary, respectively:
\begin{equation}
\centering
% 	\resizebox{0.90\linewidth}{!}{$
		\displaystyle
	-\frac{\partial_{\mathcal{L}_{ce}}}{\partial_{\mathbf{W}_k}}=
		\left\{
		\begin{array}{lr}
		\frac{1}{n\tau}\sum_{i=1}^{n_k}\mathbf{x}_i\\
		-\frac{1}{n\tau}\sum_{c=1}^{|\mathcal{Y}|}\sum_{j=1}^{n_c} p_k(\mathbf{x}_j)\mathbf{x}_j\\
		\end{array}.
		\right.
% 		$}
        \label{eqa:prop1}
	\end{equation}
\end{theorem}
To prove the proposition, we will partition the training set based on whether each sample belongs to class $k$. The gradient generated by sample $\mathbf{x}_i$ in class $k$ with respect to $\mathbf{W}_k$ is given by:
\begin{equation}
\centering
% 	\resizebox{0.90\linewidth}{!}{$
		\displaystyle
	-\frac{\partial_{\mathcal{L}_{ce}^i}}{\partial_{\mathbf{W}_k}}=\frac{1}{n\tau}[\mathbf{x}_i-p_k(\mathbf{x}_i)\mathbf{x}_i] ,
% 		$}
        \label{in}
	\end{equation}
 whereas the gradient produced by sample $\mathbf{x}_j$ outside of class $k$ with respect to $\mathbf{W}_k$ is:
\begin{equation}
\centering
% 	\resizebox{0.90\linewidth}{!}{$
		\displaystyle
	-\frac{\partial_{\mathcal{L}_{ce}^j}}{\partial_{\mathbf{W}_k}}=\frac{1}{n\tau}[-p_k(\mathbf{x}_j)\mathbf{x}_j].
% 		$}
        \label{out}
	\end{equation}
  By combining these gradients across all samples, we obtain:
\begin{equation}
\centering
% 	\resizebox{0.90\linewidth}{!}{$
		\displaystyle
	-\frac{\partial_{\mathcal{L}_{ce}}}{\partial_{\mathbf{W}_k}}=
		\frac{1}{n\tau}[\sum_{i=1}^{n_k}\mathbf{x}_i-\sum_{c=1}^{|\mathcal{Y}|}\sum_{j=1}^{n_c} p_k(\mathbf{x}_j)\mathbf{x}_j],
% 		$}
	\end{equation}
which proves proposition \ref{prop}. Eq. (\ref{eq:gradient}) in the paper is derived in the same way.
\section{PyTorch Implementation of the Revised-Cross-Entropy}
In Fig. \ref{code}, we provide our PyTorch implementation for the revised cross-entropy.
\end{document}